\documentclass{article}

    \usepackage[preprint,nonatbib]{neurips_2020}

\usepackage[utf8]{inputenc} % allow utf-8 input
\usepackage[T1]{fontenc}    % use 8-bit T1 fonts
\usepackage{hyperref}       % hyperlinks
\usepackage{url}            % simple URL typesetting
\usepackage{booktabs}       % professional-quality tables
\usepackage{amsfonts}       % blackboard math symbols
\usepackage{nicefrac}       % compact symbols for 1/2, etc.
\usepackage{microtype}      % microtypography

\usepackage{amsmath,amssymb}
\usepackage{subcaption,graphicx}
\usepackage{xcolor}
\usepackage{multirow}
\usepackage[linesnumbered,ruled]{algorithm2e}
\usepackage{wrapfig}
\usepackage{stackengine}

\title{GOLD-NAS: Gradual, One-Level, Differentiable}

\author{%
Kaifeng Bi\textsuperscript{1,2},\quad Lingxi Xie\textsuperscript{1},\quad Xin Chen\textsuperscript{1,3},\quad Longhui Wei\textsuperscript{1},\quad Qi Tian\textsuperscript{1}\\
\textsuperscript{1}Huawei Inc.\quad\textsuperscript{2}Tsinghua University\quad\textsuperscript{3}Tongji University\\
{\small\texttt{bkf16@mails.tsinghua.edu.cn}}\quad{\small\texttt{198808xc@gmail.com}}\\\quad{\small\texttt{1410452@tongji.edu.cn}}\quad{\small\texttt{weilonghui1@huawei.com}}\quad{\small\texttt{tian.qi1@huawei.com}}
}

\begin{document}

\maketitle

\begin{abstract}
There has been a large literature of neural architecture search, but most existing work made use of heuristic rules that largely constrained the search flexibility. In this paper, we first relax these manually designed constraints and enlarge the search space to contain more than $10^{160}$ candidates. In the new space, most existing differentiable search methods can fail dramatically. We then propose a novel algorithm named Gradual One-Level Differentiable Neural Architecture Search (\textbf{GOLD-NAS}) which introduces a variable resource constraint to one-level optimization so that the weak operators are gradually pruned out from the super-network. In standard image classification benchmarks, GOLD-NAS can find a series of Pareto-optimal architectures within a single search procedure. Most of the discovered architectures were never studied before, yet they achieve a nice tradeoff between recognition accuracy and model complexity. We believe the new space and search algorithm can advance the search of differentiable NAS.
\end{abstract}

\section{Introduction}
\label{introduction}

With the rapid development of deep learning~\cite{lecun2015deep}, designing powerful neural architectures has been a major challenge for the researchers in artificial intelligence. Neural architecture search (NAS)~\cite{zoph2017neural}, a recent subarea of machine learning, has attracted increasing attentions recently due to the potential of finding effective and/or efficient architectures that outperform human expertise. Existing NAS methods can be partitioned into two parts, namely, individual search methods and weight-sharing search methods. The individual search methods~\cite{zoph2018learning,real2017large,liu2018progressive} sample a large number of architectures from the search space and optimize them individually to test their performance. To reduce the computational burden, the weight-sharing search methods~\cite{cai2018efficient,pham2018efficient,liu2019darts} formulate the search space into a super-network and try to reuse computation among the sampled architectures.

This paper focuses on a special type of weight-sharing search methods named differentiable neural architecture search (DNAS~\cite{shin2018differentiable} or DARTS~\cite{liu2019darts}). These methods have the ability of jointly optimizing the network weights and architectural parameters. The final architecture is obtained after training one super-network and the search cost is reduced to several hours. DARTS has become one of the most popular NAS pipelines nowadays, but it suffers three major drawbacks listed below.
\begin{itemize}
\item The search space of DARTS is \textit{highly limited}, \textit{e.g.}, there is exactly one operator preserved for each edge, each node receives two prior inputs, \textit{etc}. These constraints are helpful for the stability of NAS, but they also shrink the accuracy gain brought by powerful search methods: with some heuristic designs (\textit{e.g.}, two \textsf{skip-connect} operators in each cell~\cite{chen2019progressive}) or search tricks (\textit{e.g.}, early termination~\cite{liang2019darts+}), even random search can achieve satisfying performance.
\item DARTS requires \textit{bi-level optimization}, \textit{i.e.}, a training phase to optimize the network weights and a validation phase to update the architecture parameters. This mechanism brings computational burden and, more importantly, considerable inaccuracy in gradient estimation that can dramatically deteriorate the search procedure~\cite{bi2019stabilizing,zela2020understanding}.
\item DARTS removes weak operators and edges all at once after the super-network has been optimized, but this step can risk a large \textit{discretization error} especially when the weights of the pruned operators are not guaranteed to be small.
\end{itemize}

To address these problems, we advocate for an enlarged search space which borrows the cell-based design of DARTS but frees most of the heuristic constraints. In particular, all cells are allowed to have different architectures, each edge can contain more than one operators, and each node can receive input from an arbitrary number of its precedents. These modifications have increased the size of search space from less than $10^{20}$ to more than $10^{160}$. More importantly, we believe the reduction of constraints will raise new challenges to the stability and advance the research of NAS methods.

In this complex space, bi-level optimization suffers from heavy computational burden as well as the inaccuracy of gradient estimation~\cite{bi2019stabilizing}. This urges us to apply one-level optimization which is easier to get rid of the computational burdens. However, as shown in~\cite{liu2019darts}, one-level optimization can run into dramatic failure which, according to our diagnosis, mainly owes to the discretization error caused by removing the moderate operators. Motivated by this finding, we present a novel framework which starts with a complete super-network and gradually prunes out weak operators. During the search procedure, we avoid applying heuristic rules but rely on resource constraints (\textit{e.g.}, FLOPs) to determine which operators should be eliminated. Our algorithm is named \textbf{GOLD-NAS} which stands for Gradual One-Level Differentiable Neural Architecture Search. Compared to prior NAS approaches, GOLD-NAS requires little human expertise and is less prone of the optimization gap.

We perform experiments on CIFAR10 and ImageNet, two popular image classification benchmarks. Within a small search cost ($0.4$ GPU-days on CIFAR10 and $1.3$ GPU-days on ImageNet), GOLD-NAS find a series of Pareto-optimal architectures that can fit into different hardware devices. In particular, the found architectures achieve a $2.99\pm0.05\%$ error on CIFAR10 with merely $1.58\mathrm{M}$ parameters, and a $23.9\%$ top-1 error on ImageNet under the mobile setting. These results pave the way of searching in a much larger space which is very challenging for most prior work.

\section{Related Work}
\label{related_work}

Neural architecture search~\cite{zoph2017neural} provides an automatic way to alleviates the burden of manually designing network architectures~\cite{krizhevsky2012imagenet,simonyan2015very,szegedy2015going,he2016deep,huang2017densely}. Popular NAS methods often start with defining the \textbf{search space} which contains all possible architectures and follow a heuristic \textbf{search algorithm} to explore the space efficiently to find the optimal architecture.

A good search space often has a large capacity so that there exist high-quality architectures (either of high quality or of high efficiency) and these architectures are difficult to find following some manually defined rules. Currently popular search spaces~\cite{zoph2018learning,liu2019darts,howard2019searching} are often composed of some repeatable \textbf{cells}, each of which is a relatively complex combination of basic operators (\textit{e.g.}, convolution, pooling, \textit{etc.}). Each cell receives inputs from previous cells, and the connectivity between these cells can be either fixed~\cite{liu2019darts,howard2019searching} or searched~\cite{real2017large,xie2017genetic}.

The very first search algorithms~\cite{zoph2017neural,real2017large,xie2017genetic} explored by the researchers involves sampling architectures from the search space, evaluating them individually, and using the evaluation results to update the heuristic function that depicts the search space. These algorithms are often slow and difficult to generalize across datasets, so later efforts focused on reusing computation of similar architectures~\cite{cai2018efficient,luo2018neural}. This path eventually leads to the \textbf{one-shot} search methods~\cite{liu2019darts,chu2019fairnas,guo2019single} that trains the super-network only once, after which the sub-networks are sampled and evaluated.

A special type of the one-shot search methods is named \textbf{differentiable search}~\cite{luo2018neural,shin2018differentiable,liu2019darts}, in which the search space is relaxed so that the architectural parameters can take continuous values and thus can be optimized together with the network weights in a gradient descent process. Since the set of architectural parameters is often much smaller than that of network weights, a relatively safe flowchart, as proposed in DARTS~\cite{liu2019darts}, is bi-level optimization, \textit{i.e.}, partitioning each search step into two phases for updating the network weights and architectural parameters, respectively. However, bi-level optimization also brings considerable inaccuracy to gradient estimation~\cite{bi2019stabilizing} which reflects as the instability of the search process, \textit{e.g.}, the searched cells collapse to dummy ones. Researchers proposed heuristic optimization tricks that work well in constrained cases~\cite{chen2019progressive,liang2019darts+,zela2020understanding}, but these tricks can still fail when the search space continues to expand~\cite{bi2019stabilizing}. One-level optimization was investigated but believed difficult~\cite{liu2019darts} unless the search space is modified~\cite{li2019stacnas}.

\section{Building GOLD-NAS on CIFAR10}
\label{cifar}

\subsection{Differentiable NAS: Dataset, Settings, and Overview}
\label{cifar:overview}

The CIFAR10 dataset~\cite{krizhevsky2009learning} is one of the most popular benchmarks for neural architecture search. It has $50\mathrm{K}$ training images and $10\mathrm{K}$ testing images, both of which uniformly distributed over $10$ classes. Each image is RGB and has a resolution of $32\times32$. We follow the convention to first determine the optimal architecture and then re-train it for evaluation. The test set remains invisible in both the search and re-training phases. Detailed hyper-parameter settings are elaborated in Appendix~\ref{cifar_:parameters}.

We start with defining an enlarged search space (Section~\ref{cifar:space}) that discards most manual designs and provides a better benchmark for NAS evaluation. Next, we demonstrate the need of one-level optimization (Section~\ref{cifar:onelevel}) and analyze the difficulty of performing discretization in the new space (Section~\ref{cifar:discretization}). Finally, to solve the problem, we design a pruning algorithm (Section~\ref{cifar:pruning}) that gradually eliminates weak operators and/or edges with the regularization of resource efficiency.

\subsection{Breaking the Rules: Enlarging the Search Space}
\label{cifar:space}

We first recall the cell-based super-network used in DARTS~\cite{liu2019darts}. It has a fixed number ($L$) of cells. Each cell has two input signals from the previous two cells (denoted as $\mathbf{x}_0$ and $\mathbf{x}_1$), and $N-2$ inner nodes to store intermediate responses. For each ${i}<{j}$ except for ${\left(i,j\right)}={\left(0,1\right)}$, the output of the $i$-th node is sent to the $j$-th node via the edge of $\left(i,j\right)$. Mathematically, we have ${g_{i,j}\left(\mathbf{x}_i\right)}={{\sum_{o\in\mathcal{O}}}\sigma\!\left(\alpha_{i,j}^o\right)\cdot o\!\left(\mathbf{x}_i\right)}$ where $\mathbf{x}_i$ is the output of the $i$-th node, $\mathcal{O}$ is a pre-defined set of operators, and $o\!\left(\cdot\right)$ is an element in $\mathcal{O}$. $\sigma\!\left(\alpha_{i,j}^o\right)$ determines the weight of $o\!\left(\mathbf{x}_i\right)$, which is set to be ${\sigma\!\left(\alpha_{i,j}^o\right)}={\exp(\alpha_{i,j}^o)/{\sum_{o'\in\mathcal{O}}}\exp(\alpha_{i,j}^{o'})}$. The output of the $j$-th cell is the sum of all information flows from the precursors, \textit{i.e.}, ${\mathbf{x}_j}={\sum_{i<j}g_{i,j}\left(\mathbf{x}_i\right)}$, and the final output of the cell is the concatenation of all non-input nodes, \textit{i.e.}, $\mathrm{concat}\!\left(\mathbf{x}_2,\mathbf{x}_3,\ldots,\mathbf{x}_{N-1}\right)$. In this way, the super-network is formulated into a differentiable function, ${f\!\left(\mathbf{x}\right)}\doteq{f\!\left(\mathbf{x};\boldsymbol{\alpha},\boldsymbol{\omega}\right)}$, where $\boldsymbol{\alpha}$ and $\boldsymbol{\omega}$ indicate the architectural parameters and network weights, respectively.

DARTS~\cite{liu2019darts} and its variants~\cite{chen2019progressive,nayman2019xnas,xu2020pc} have relied on many manually designed rules to determine the final architecture. Examples include each edge can only preserve one operator, each inner node can preserve two of its precursors, and the architecture is shared by the same type (normal and reduction) of cells. These constraints are helpful for the stability of the search process, but they limit the flexibility of architecture search, \textit{e.g.}, the low-level layers and high-level layers must have the same topological complexity which is no reason to be the optimal solution. A prior work~\cite{bi2019stabilizing} delivered an important message that the ability of NAS approaches is better evaluated in a more complex search space (in which very few heuristic rules are used). Motivated by this, we release the heuristic constraints to offer higher flexibility to the final architecture, namely, each edge can preserve an arbitrary number of operators (they are directly summed into the output), each inner node can preserve an arbitrary number of precedents, and all cell architectures are independent.

To fit the new space, we slightly modify the super-network so that $\sigma\!\left(\alpha_{i,j}^o\right)$ is changed from the softmax function to element-wise sigmoid, \textit{i.e.}, ${\sigma\!\left(\alpha_{i,j}^o\right)}={\exp(\alpha_{i,j}^o)/\left(1+\exp(\alpha_{i,j}^o)\right)}$. This offers a more reasonable basis to the search algorithm since the enhancement of any operator does not necessarily lead to the attenuation of all others~\cite{chu2019fair}. Moreover, the independence of all cells raises the need of optimizing the complete super-network (\textit{e.g.}, having $20$ cells) during the search procedure. To fit the limited GPU memory, we follow~\cite{bi2019stabilizing,zela2020understanding} to preserve two operators, \textsf{skip-connect} and \textsf{sep-conv-3x3}, in each edge. Note that the reduction of candidate operators does not mean the search task has become easier. Even with two candidates per edge, the new space contains as many as $6.9\times10^{167}$ architectures\footnote{This is the theoretical maximum. Under the resource constraints (Section~\ref{cifar:pruning}), the final architecture is often in a relatively small space, but the space is still much larger than the competitors. Please refer to Appendix~\ref{details:space} for the calculation of the number of possible architectures.}, which is far more than the capacity of the original space with either shared cells ($1.1\times10^{18}$,~\cite{liu2019darts}) or individual cells ($1.9\times10^{93}$,~\cite{bi2019stabilizing}). Without heuristic rules, exploring this enlarged space requires more powerful search methods.

\subsection{Why One-Level Optimization?}
\label{cifar:onelevel}

The goal of differentiable NAS is to solve the following optimization:
\begin{equation}
\label{eqn:goal1}
{\boldsymbol{\alpha}^\star}={\arg\min_{\boldsymbol{\alpha}}\mathcal{L}\!\left(\boldsymbol{\omega}^\star\!\left(\alpha\right),\boldsymbol{\alpha};\mathcal{D}_\mathrm{train}\right)},\quad\quad\mathrm{s.t.}\quad{\boldsymbol{\omega}^\star\!\left(\alpha\right)}={\arg\min_{\boldsymbol{\omega}}\mathcal{L}\!\left(\boldsymbol{\omega},\boldsymbol{\alpha};\mathcal{D}_\mathrm{train}\right)},
\end{equation}
where ${\mathcal{L}\!\left(\boldsymbol{\omega},\boldsymbol{\alpha};\mathcal{D}_\mathrm{train}\right)}={\mathbb{E}_{\left(\mathbf{x},\mathbf{y}^\star\right)\in\mathcal{D}_\mathrm{train}}\!\left[\mathrm{CE}\!\left(f\!\left(\mathbf{x}\right),\mathbf{y}^\star\right)\right]}$ is the loss function computed in a specified training dataset. There are mainly two methods for this purpose, known as one-level optimization and bi-level (two-level) optimization, respectively. Starting with $\boldsymbol{\alpha}_0$ and $\boldsymbol{\omega}_0$, the initialization of $\boldsymbol{\alpha}$ and $\boldsymbol{\omega}$, \textit{one-level optimization} involves updating $\boldsymbol{\alpha}$ and $\boldsymbol{\omega}$ simultaneously in each step:
\begin{equation}
\label{eqn:onelevel}
{\boldsymbol{\omega}_{t+1}}\leftarrow{\boldsymbol{\omega}_t-\eta_{\boldsymbol{\omega}}\cdot\nabla_{\boldsymbol{\omega}}\mathcal{L}\!\left(\boldsymbol{\omega}_t,\boldsymbol{\alpha}_t;\mathcal{D}_\mathrm{train}\right)},\quad\quad{\boldsymbol{\alpha}_{t+1}}\leftarrow{\boldsymbol{\alpha}_t-\eta_{\boldsymbol{\alpha}}\cdot\nabla_{\boldsymbol{\alpha}}\mathcal{L}\!\left(\boldsymbol{\omega}_t,\boldsymbol{\alpha}_t;\mathcal{D}_\mathrm{train}\right)}.
\end{equation}
Note that, since the numbers of parameters in $\boldsymbol{\alpha}$ and $\boldsymbol{\omega}$ differ significantly (from tens to millions), different learning rates ($\eta_{\boldsymbol{\alpha}}$ and $\eta_{\boldsymbol{\omega}}$) and potentially different optimizers can be used. Even in this way, the algorithm is easily biased towards optimizing $\boldsymbol{\omega}$, leading to unsatisfying performance\footnote{This is because of the imbalanced effect brought by optimizing $\boldsymbol{\alpha}$ and $\boldsymbol{\omega}$, in which the latter is often more effective. An intuitive example is to fix $\boldsymbol{\alpha}$ as the status after random initialization and only optimize $\boldsymbol{\omega}$, which can still leads to a high accuracy in the training data (it is not possible to achieve this goal by only optimizing $\boldsymbol{\alpha}$). Obviously, this does not deliver any useful information to architecture design.}. To fix this issue, a practical way is to evaluate the performance with respect to $\boldsymbol{\alpha}$ and $\boldsymbol{\omega}$ in two separate training sets, \textit{i.e.}, ${\mathcal{D}_\mathrm{train}}={\mathcal{D}_1\cup\mathcal{D}_2}$. Hence, the goal of optimization becomes:
\begin{equation}
\label{eqn:goal2}
{\boldsymbol{\alpha}^\star}={\arg\min_{\boldsymbol{\alpha}}\mathcal{L}\!\left(\boldsymbol{\omega}^\star\!\left(\alpha\right),\boldsymbol{\alpha};\mathcal{D}_1\right)},\quad\quad\mathrm{s.t.}\quad{\boldsymbol{\omega}^\star\!\left(\alpha\right)}={\arg\min_{\boldsymbol{\omega}}\mathcal{L}\!\left(\boldsymbol{\omega},\boldsymbol{\alpha};\mathcal{D}_2\right)},
\end{equation}
and, correspondingly, \textit{bi-level optimization} is used to update $\boldsymbol{\alpha}$ and $\boldsymbol{\omega}$ alternately:
\begin{equation}
\label{eqn:bilevel}
{\boldsymbol{\omega}_{t+1}}\leftarrow{\boldsymbol{\omega}_t-\eta_{\boldsymbol{\omega}}\cdot\nabla_{\boldsymbol{\omega}}\mathcal{L}\!\left(\boldsymbol{\omega}_t,\boldsymbol{\alpha}_t;\mathcal{D}_2\right)},\quad\quad{\boldsymbol{\alpha}_{t+1}}\leftarrow{\boldsymbol{\alpha}_t-\eta_{\boldsymbol{\alpha}}\cdot\nabla_{\boldsymbol{\alpha}}\mathcal{L}\!\left(\boldsymbol{\omega}_{t+1},\boldsymbol{\alpha}_t;\mathcal{D}_1\right)}.
\end{equation}
DARTS~\cite{liu2019darts} tried both optimization methods and advocated for the superiority of bi-level optimization. However, as pointed out in~\cite{bi2019stabilizing}, bi-level optimization suffers considerable inaccuracy of gradient estimation and the potential instability can increase with the complexity of the search space. This drives us back to one-level optimization. Fortunately, we find that the failure of one-level optimization can be easily prevented. Detailed analyses and experiments are provided in Appendix~\ref{details:onelevel}. Here, we deliver the key message that one-level optimization is made quite stable by adding regularization (\textit{e.g.}, Cutout~\cite{devries2017improved}, AutoAugment~\cite{cubuk2019autoaugment}, \textit{etc.}) to a small dataset (\textit{e.g.}, CIFAR10) or simply using a large dataset (\textit{e.g.}, ImageNet). So, we focus on applying one-level optimization to the enlarged search space throughout the remaining part of this paper.

\subsection{The Difficulty of Discretization}
\label{cifar:discretization}

The main challenge that we encounter in the enlarged space is the difficulty of performing discretization, \textit{i.e.}, determining the final architecture based on $\boldsymbol{\alpha}$. This is to require $\boldsymbol{\alpha}$ in Eqn~\eqref{eqn:goal1} to satisfy the condition that ${\sigma\!\left(\alpha_{i,j}^o\right)}={\exp(\alpha_{i,j}^o)/\left(1+\exp(\alpha_{i,j}^o)\right)}$ is very close to $0$ or $1$, but this constraint is difficult to be integrated into a regular optimization process like Eqn~\eqref{eqn:onelevel}. The solution of conventional approaches~\cite{liu2019darts,chen2019progressive,xu2020pc,zela2020understanding} is to perform hard pruning at the end of the search stage to eliminate weak operators from the super-network, \textit{e.g.}, an operator is preserved if ${\sigma\!\left(\alpha_{i,j}^o\right)}>{0.5}$.

This algorithm can lead to significant \textit{discretization error}, since many of the pruned operators have moderate weights, \textit{i.e.}, ${\sigma\!\left(\alpha_{i,j}^o\right)}={\exp(\alpha_{i,j}^o)/\left(1+\exp(\alpha_{i,j}^o)\right)}$ is neither close to $0$ nor $1$. In this scenario, directly removing these operators can lead to dramatic accuracy drop on the super-network. Mathematically, this may push $\boldsymbol{\alpha}$ (and also $\boldsymbol{\omega}\!\left(\boldsymbol{\alpha}\right)$) away from the current optimum, so that the algorithm may need a long training process to arrive at another optimum, or never. The reason for $\sigma\!\left(\alpha_{i,j}^o\right)$ being moderate is straightforward: the new space allows an arbitrary number of operators to be preserved on each edge, or more specifically, there is no internal mechanism for the operators to compete with each other. Therefore, the best strategy to fit training data is to keep \textbf{all} the operators, since almost all operators contribute more or less to the training accuracy, but this is of little use to architecture search itself.

Motivated by this, we propose to add regularization to the process of super-network training so that to penalize the architectures that use more computational resources. This mechanism is similar in a few prior work that incorporated hardware constraints to the search algorithm~\cite{cai2019proxylessnas,tan2019mnasnet,wu2019fbnet}, but the goal of our method is to use the power of regularization to suppress the weight of some operators so that they can be pruned. Note that the risk of discretization error grows as the number and the strength (weights) of pruned operators. So, a safe choice is to perform pruning multiple times, in each of which only the operators with sufficiently low weights can be removed. We elaborate the details in the following subsection.

\begin{algorithm}[!b]
\SetKwInOut{Input}{Input}
\SetKwInOut{Output}{Output}
\SetKwInOut{Return}{Return}
\Input{
Search space $\mathcal{S}$, dataset $\mathcal{D}_\mathrm{train}$, balancing coefficient $\mu$, minimal FLOPs constraints $\mathrm{FLOPs}_\mathrm{min}$, learning rates $\eta_{\boldsymbol{\omega}}$, $\eta_{\boldsymbol{\alpha}}$, pruning hyper-parameters $n_0$, $\lambda_0$, $c_0$, $\xi_\mathrm{max}$, $\xi_\mathrm{min}$, $t_0$;
}
\Output{
A set of pareto-optimal architectural parameters $\mathcal{A}$;
}
Initialize $\boldsymbol{\omega}^\mathrm{curr}$ and $\boldsymbol{\alpha}^\mathrm{curr}$ as random noise, ${\mathcal{A}}\leftarrow{\varnothing}$, ${\lambda}\leftarrow{0}$, $\Delta{\lambda}\leftarrow{\lambda_0}$, ${t}\leftarrow{0}$;\\
\Repeat{${\mathrm{FLOPs}\!\left(\boldsymbol{\alpha}^\mathrm{curr}\right)}\leqslant{\mathrm{FLOPs}_\mathrm{min}}$}{
Update $\boldsymbol{\omega}^\mathrm{curr}$ and $\boldsymbol{\alpha}^\mathrm{curr}$ using Eqn~\eqref{eqn:onelevel} for one epoch;\\
Let $\mathcal{E}$ be the set of active operators, and $\mathcal{E}_\mathrm{min}$ be the $n_0$ operators in $\mathcal{E}$ with minimal weights;\\
Prune operators in $\mathcal{E}_\mathrm{min}$ with weight smaller than $\xi_\mathrm{max}$, and operators in $\mathcal{E}$ with weight smaller than $\xi_\mathrm{min}$, let $n_\mathrm{pruned}$ be the number of pruned operators;\\
\textbf{if} ${n_\mathrm{pruned}}<{n_0}$ \textbf{then} $\Delta\lambda\leftarrow{c_0\Delta\lambda}$, ${\lambda}\leftarrow{\lambda+\Delta\lambda}$ \textbf{else} $\Delta\lambda\leftarrow{\lambda_0}$, ${\lambda}\leftarrow{\lambda/c_0}$;\\
\textbf{if} ${n_\mathrm{pruned}}={0}$ \textbf{then} ${t}\leftarrow{t+1}$ \textbf{else} ${t}\leftarrow{0}$;\\
\textbf{if} ${t}\geqslant{t_0}$ \textbf{then} ${\mathcal{A}}\leftarrow{\mathcal{A}\cup\left\{\boldsymbol{\alpha}^\mathrm{curr}\right\}}$, ${t}\leftarrow{0}$;\\
}
\Return{
$\mathcal{A}$.
}
\caption{
Gradual One-Level Differentiable Neural Architecture Search (\textbf{GOLD-NAS})
}
\label{alg:pipeline}
\end{algorithm}

\subsection{Gradual Pruning with Resource Constraints}
\label{cifar:pruning}

To satisfy the condition that the weights of pruned operators are sufficiently small, we design a gradual pruning algorithm. The core idea is to start with a low regularization coefficient and increase it gradually during the search procedure. Every time the coefficient becomes larger, there will be some operators (those having higher redundancy) being suppressed to low weights. Pruning them out causes little drop in training accuracy. This process continues till the super-network becomes sufficiently small. During the search process, The architectures that survive for sufficiently long are recorded, which compose the set of Pareto-optimal architectures.

Throughout the remaining part, we set the regularization term as the expected FLOPs of the super-network, and this framework can be generalized to other kinds of constraints (\textit{e.g.}, network latency~\cite{wu2019fbnet,tan2019mnasnet,xu2020latency}). Conceptually, adding resource constraints requires a slight modification to the objective function, Eqn~\eqref{eqn:goal1}. With $\mathrm{FLOPs}\!\left(\boldsymbol{\alpha}\right)$ denoting the expected FLOPs under the architectural parameter of $\boldsymbol{\alpha}$, the overall optimization goal is to achieve a tradeoff between accuracy and efficiency. We have carefully designed the calculation of $\mathrm{FLOPs}\!\left(\boldsymbol{\alpha}\right)$, described in Appendix~\ref{details:flops}, so that (i) the result strictly equals to the evaluation of the \texttt{thop} library, and (ii) $\mathrm{FLOPs}\!\left(\boldsymbol{\alpha}\right)$ is differentiable to $\boldsymbol{\alpha}$. 

The top-level design of gradual pruning is to facilitate the competition between accuracy and resource efficiency. For this purpose, we modify the original objective function to incorporate the FLOPs constraint, \textit{i.e.}, ${\mathcal{L}\!\left(\boldsymbol{\omega},\boldsymbol{\alpha}\right)}={\mathbb{E}_{\left(\mathbf{x},\mathbf{y}^\star\right)\in\mathcal{D}_\mathrm{train}}\!\left[\mathrm{CE}\!\left(f\!\left(\mathbf{x}\right),\mathbf{y}^\star\right)\right]+\lambda\cdot\left(\overline{\mathrm{FLOPs}}\!\left(\boldsymbol{\alpha}\right)+\mu\cdot\mathrm{FLOPs}\!\left(\boldsymbol{\alpha}\right)\right)}$, where the two coefficients, $\lambda$ and $\mu$, play different roles. $\lambda$ starts with $0$ and vibrates during the search procedure to smooth the pruning process, resulting in a Pareto front that contains a series of optimal architectures with different computational costs. $\mu$ balances between $\mathrm{FLOPs}\!\left(\boldsymbol{\alpha}\right)$ and $\overline{\mathrm{FLOPs}}\!\left(\boldsymbol{\alpha}\right)$, the \textit{expected} and \textit{uniform} versions of FLOPs calculation (see Appendix~\ref{details:flops}). In brief, $\mathrm{FLOPs}\!\left(\boldsymbol{\alpha}\right)$ adds lower penalty to the operators with smaller computational costs, but $\overline{\mathrm{FLOPs}}\!\left(\boldsymbol{\alpha}\right)$ adds a fixed weight to all operators. Hence, a larger $\mu$ favors pruning more convolutions and often pushes the architecture towards higher computational efficiency. In other words, one can tune the value of $\mu$ to achieve different Pareto fronts (see the later experiments).

The overall search procedure is summarized in Algorithm~\ref{alg:pipeline}. At the beginning, $\lambda$ is set to be $0$ and the training procedure focuses on improving accuracy. As the optimization continues, $\lambda$ gradually goes up and forces the network to reduce the weight on the operators that have fewer contribution. In each pruning round, there is an expected number of operators to be pruned. If this amount is not achieved, $\lambda$ continues to increase, otherwise it is reduced. If no operators are pruned for a few consecutive epochs, the current architecture is considered Pareto-optimal and added to the output set. Our algorithm is named \textbf{GOLD-NAS}, which indicates its most important properties: \textbf{G}radual, \textbf{O}ne-\textbf{L}evel, and \textbf{D}ifferentiable. \textcolor{blue}{We emphasize that it is the \textbf{gradual pruning} strategy that alleviates the discretization error and enables the algorithm to benefit from the flexibility of \textbf{one-level optimization} and the efficiency of \textbf{differentiable search}.}

\begin{table}[!b]
\begin{center}
\setlength{\tabcolsep}{0.16cm}
\begin{tabular}{lccccccccc}
\hline
\multicolumn{2}{l}{\textbf{\multirow{2}{*}{Architecture}}} & \multicolumn{4}{c}{\textbf{Test Err. (\%)}} & \textbf{Params} & \textbf{Search Cost} & \textbf{FLOPs} \\
\cmidrule(lr){3-6}
{} & {} & \textbf{\#1} & \textbf{\#2} & \textbf{\#3} & \textbf{average} & \textbf{(M)} & \textbf{(GPU-days)} & \textbf{(M)} \\
\hline
\multicolumn{2}{l}{DenseNet-BC~\cite{huang2017densely}}            &      &      &      & 3.46          & 25.6 & -           & - \\
\hline
\multicolumn{2}{l}{ENAS~\cite{pham2018efficient}}                  &      &      &      & 2.89          & 4.6  & 0.5         & 626 \\
\multicolumn{2}{l}{NASNet-A~\cite{zoph2018learning}}               &      &      &      & 2.65          & 3.3  & 1800        & 605 \\
\multicolumn{2}{l}{AmoebaNet-B~\cite{real2019regularized}}         &      &      &      & 2.55$\pm$0.05 & 2.8  & 3150        & 490 \\
\hline
\multicolumn{2}{l}{SNAS (moderate)~\cite{xie2018snas}}             &      &      &      & 2.85$\pm$0.02 & 2.8  & 1.5         & 441 \\
\multicolumn{2}{l}{DARTS (\textit{1st-order})~\cite{liu2019darts}} &      &      &      & 3.00$\pm$0.14 & 3.3  & 0.4         & - \\
\multicolumn{2}{l}{DARTS (\textit{2nd-order})~\cite{liu2019darts}} &      &      &      & 2.76$\pm$0.09 & 3.3  & 1.0         & 528 \\
\multicolumn{2}{l}{P-DARTS~\cite{chen2019progressive}}             &      &      &      & 2.50          & 3.4  & 0.3         & 532 \\
\multicolumn{2}{l}{PC-DARTS~\cite{xu2020pc}}                       &      &      &      & 2.57$\pm$0.07 & 3.6  & 0.1         & 557 \\
\hline
% \multicolumn{2}{l}{Random search baseline}                       &      &      &      & 3.31$\pm$0.50 & 2.3$\pm$0.5  & 4.0         & 368$\pm73$ \\
% \hline
GOLD-NAS-A & \multirow{6}{*}{${\mu}={1}$}                          & 2.93 & 3.02 & 3.01 & 2.99$\pm$0.05 & 1.58 & 0.4         & 245 \\
GOLD-NAS-B &                                                       & 2.97 & 2.85 & 3.08 & 2.97$\pm$0.12 & 1.72 & 0.4         & 267 \\
GOLD-NAS-C &                                                       & 2.94 & 2.97 & 2.97 & 2.96$\pm$0.02 & 1.76 & 0.4         & 287 \\
GOLD-NAS-D &                                                       & 2.89 & 2.98 & 2.84 & 2.90$\pm$0.07 & 1.89 & 0.4         & 308 \\
GOLD-NAS-E &                                                       & 2.75 & 2.86 & 2.89 & 2.83$\pm$0.07 & 1.99 & 0.4         & 334 \\
GOLD-NAS-F &                                                       & 2.77 & 2.79 & 2.86 & 2.81$\pm$0.05 & 2.08 & 0.4         & 355 \\
\hline
GOLD-NAS-G & \multirow{6}{*}{${\mu}={0}$}                          & 2.73 & 2.84 & 2.67 & 2.75$\pm$0.09 & 2.22 & 1.1         & 376 \\
GOLD-NAS-H &                                                       & 2.71 & 2.76 & 2.62 & 2.70$\pm$0.07 & 2.51 & 1.1         & 402 \\
GOLD-NAS-I &                                                       & 2.52 & 2.72 & 2.60 & 2.61$\pm$0.10 & 2.85 & 1.1         & 445 \\
GOLD-NAS-J &                                                       & 2.53 & 2.67 & 2.60 & 2.60$\pm$0.07 & 3.01 & 1.1         & 459 \\
GOLD-NAS-K &                                                       & 2.67 & 2.40 & 2.65 & 2.57$\pm$0.15 & 3.30 & 1.1         & 508 \\
GOLD-NAS-L &                                                       & 2.57 & 2.58 & 2.44 & 2.53$\pm$0.08 & 3.67 & 1.1         & 546 \\
\hline
\end{tabular}
\end{center}
\caption{Comparison to state-of-the-art NAS methods on CIFAR10. The architectures A--F are the Pareto-optimal obtained from a single search procedure (${\mu}={1}$, better efficiency), and G--L from another procedure (${\mu}={0}$). The search cost is for all six architectures sharing the same $\mu$. Each searched architecture is re-trained three times individually.}
\label{tab:comparisonC10}
\end{table}

\subsection{Results and Comparison to Prior Work}
\label{cifar:comparison}

\begin{wrapfigure}{r}{6.0cm}
\centering
\vspace{-0.4cm}
\includegraphics[width=\linewidth]{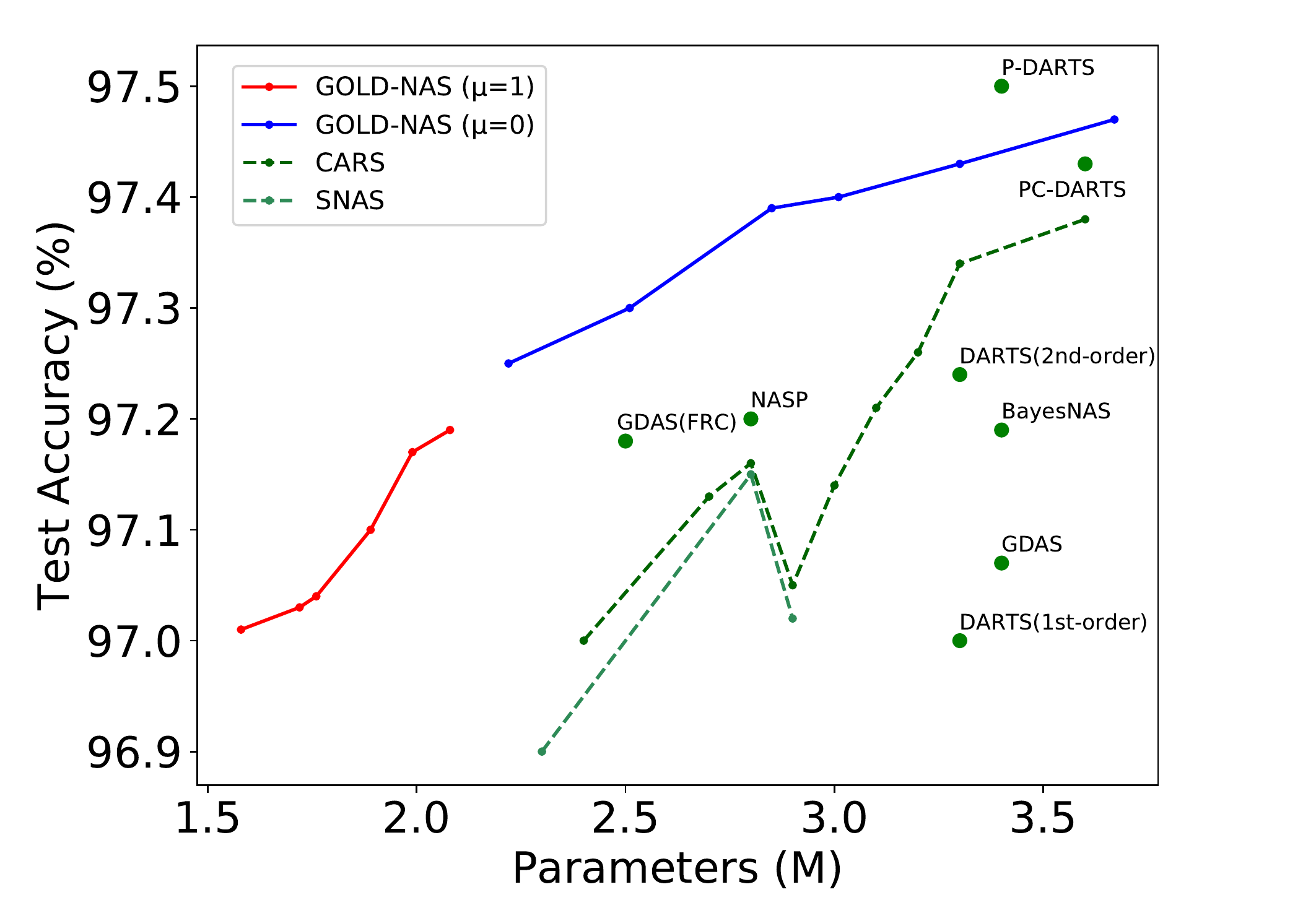}
\caption{The accuracy-complexity tradeoff of differentiable search methods. The two Pareto-fronts obtained by GOLD-NAS are shown in \textcolor{red}{red} (${\mu}={1}$) and \textcolor{blue}{blue} (${\mu}={0}$), respectively.}
\vspace{-0.2cm}
\label{fig:tradeoff}
\end{wrapfigure}

A notable benefit of GOLD-NAS is to obtain a set of Pareto-optimal architectures within one search procedure (on CIFAR10, around $10$ hours on a single NVIDIA Tesla V100 card). We use two sparsity coefficients, ${\mu}={1}$ and ${\mu}={0}$, and obtain six architectures for each, with FLOPs varying from $245\mathrm{M}$ to $546\mathrm{M}$. The intermediate results of the search procedure (\textit{e.g.}, how the values of $\lambda$ and $\left\{\sigma\!\left(\alpha_{i,j}^o\right)\right\}$ change with epochs) are shown in Appendix~\ref{cifar_:procedure}. We re-train each architecture three times, and the results are shown in Table~\ref{tab:comparisonC10}. One can observe the tradeoff between accuracy and efficiency. In particular, the GOLD-NAS-A architecture, with only $1.58\mathrm{M}$ parameters and $245\mathrm{M}$ FLOPs, achieves a $2.99\%$ error on the CIFAR10 test set. To the best of our knowledge, it is the most light-weighted model to beat the $3\%$-error mark.

We visualize the first and last architectures obtained from ${\mu}={0}$ and ${\mu}={1}$ in Figure~\ref{fig:visualizationC10}, and complete results are provided in Appendix~\ref{cifar_:visualization}. Moreover, compared to the architectures found in the original DARTS space, our algorithm allows the resource to be flexibly assigned to different stages (\textit{e.g.}, the cells close to the output does not need much resource), and this is the reason for being more efficient. From this perspective, the enlarged search space creates more opportunities for the NAS approach, yet it is the stability of GOLD-NAS that eases the exploration in this space without heuristic rules.

\begin{figure}[!t]
\centering
\begin{subfigure}{6.6cm}
\centering
\includegraphics[width=6.5cm]{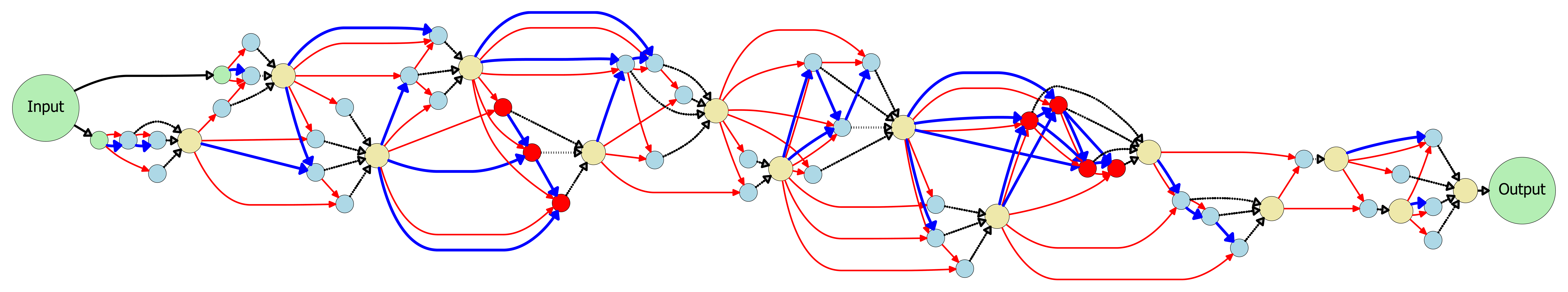}
\caption{GOLD-NAS-A, $1.58\mathrm{M}$, $2.99\%$ error}
\end{subfigure}
\hfill
\begin{subfigure}{6.6cm}
\centering
\includegraphics[width=6.5cm]{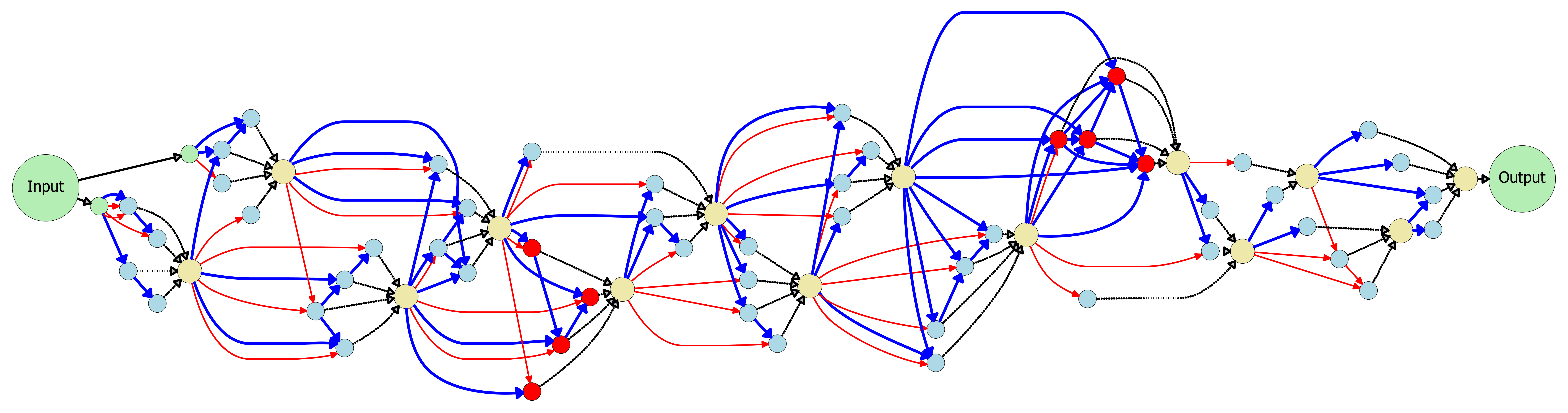}
\caption{GOLD-NAS-G, $2.22\mathrm{M}$, $2.75\%$ error}
\end{subfigure}
\begin{subfigure}{6.6cm}
\centering
\includegraphics[width=6.5cm]{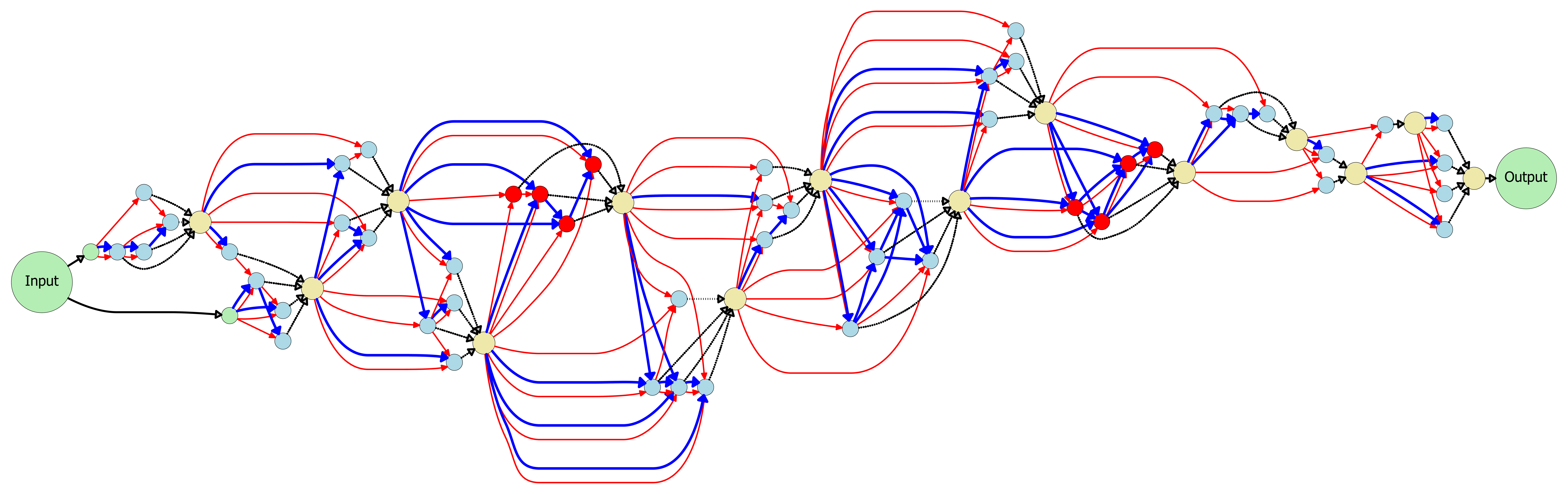}
\caption{GOLD-NAS-F, $2.08\mathrm{M}$, $2.81\%$ error}
\end{subfigure}
\hfill
\begin{subfigure}{6.6cm}
\vspace{0.5cm}
\centering
\includegraphics[width=6.5cm]{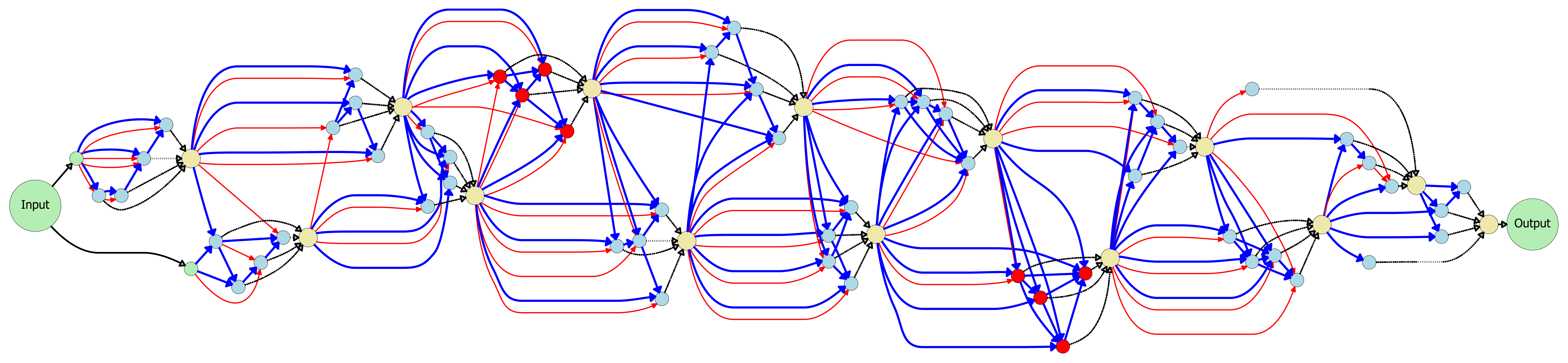}
\caption{GOLD-NAS-L, $3.67\mathrm{M}$, $2.53\%$ error}
\end{subfigure}
\caption{The first (with the highest efficiency) and last (with the highest accuracy) architectures found by two search procedures with ${\mu}={1}$ (left) and ${\mu}={0}$ (right). The \textcolor{red}{red} thin, \textcolor{blue}{blue} thick, and black dashed arrows indicate \textsf{skip-connect}, \textsf{sep-conv-3x3}, and \textsf{concatenation}, respectively. This figure is best viewed in a colored and zoomed-in document.}
\label{fig:visualizationC10}
\end{figure}

We compare our approach to the state-of-the-arts in Table~\ref{tab:comparisonC10}. Besides the competitive performance, we claim three major advantages. \textbf{First, GOLD-NAS is faster, easier to implement, and more stable than most DARTS-based methods.} This is mainly because bi-level optimization requires strict mathematical conditions ($\boldsymbol{\omega}$ needs to be optimal when $\boldsymbol{\alpha}$ gets updated, which is almost impossible to guarantee~\cite{liu2019darts}), yet the second-order gradient is very difficult to be accurately estimated~\cite{bi2019stabilizing}. In comparison, GOLD-NAS is built on one-level optimization and avoids these burdens. Meanwhile, some useful/efficient optimization methods (\textit{e.g.}, partial channel connection~\cite{xu2020pc}) can be incorporated into GOLD-NAS towards better search performance. \textbf{Second, GOLD-NAS achieves better tradeoff between accuracy and efficiency,} as shown in Figure~\ref{fig:tradeoff}. This mainly owes to its flexibility of assigning computational resources. \textbf{Third, GOLD-NAS finds a set of Pareto-optimal architectures within one search procedure.} This is more efficient than existing methods that achieved the same goal running individual search procedures with different constraints~\cite{xie2018snas} or coefficients~\cite{xu2020latency}.

Last but not least, we investigate the performance of random search. Following prior work~\cite{li2019random,liu2019darts}, we individually sample $24$ valid architectures from the new space and evaluate the performance in a $100$-epoch validation process (for technical details, please refer to Appendix~\ref{cifar_:random}). The best architecture is taken into a standard re-training process. We perform random search three times, each of which takes $4$ GPU-days, and report an average error of $3.31\pm0.50\%$, number of parameters of $2.30\pm0.49\mathrm{M}$, and FLOPs of $368\pm73\mathrm{M}$. This is far behind the Pareto fronts shown in Figure~\ref{fig:tradeoff}, indicating the strong ability of GOLD-NAS in finding efficient architectures.

\section{Generalizing GOLD-NAS to ImageNet}
\label{imagenet}

To reveal the generalization ability, we evaluate GOLD-NAS on the ImageNet-1K (ILSVRC2012) dataset~\cite{deng2009imagenet,russakovsky2015imagenet}, which contains $1.3\mathrm{M}$ training and $50\mathrm{K}$ testing images. Following~\cite{xu2020pc}, we both transfer the searched architectures from CIFAR and directly search for architectures on ImageNet. The search space remains unchanged as in CIFAR10, but three convolution layers of a stride of $2$ are inserted between the input image and the first cell, down-sampling the image size from $224\times224$ to $28\times28$. Other hyper-parameter settings are mostly borrowed from~\cite{xu2020pc}, as described in Appendix~\ref{imagenet_:parameters}.

A common protocol of ImageNet-1K is to compete under the mobile setting, \textit{i.e.}, the FLOPs of the searched architecture does not exceed $600\mathrm{M}$. We perform three individual search procedures with the basic channel number being $44$, $46$, and $48$, respectively. We set ${\mu}={0}$ to achieve higher accuracy. From each Pareto front, we take the architecture that has the largest (but smaller than $600\mathrm{M}$) FLOPs for re-training. The three architectures with basic channels numbers of $44$, $46$ and $48$ are assigned with code of X--Z, respectively. We also transplant a smaller architecture (around $500\mathrm{M}$ FLOPs) found in the $44$-channel search process to ($590\mathrm{M}$ FLOPs) by increasing the channel number from $44$ to $48$ -- we denote this architecture as GOLD-NAS-Z-tr.

\begin{wraptable}{r}{9.6cm}
\begin{center}
\setlength{\tabcolsep}{0.04cm}
\begin{tabular}{lccccc}
\hline
\textbf{\multirow{2}{*}{Architecture}} & \multicolumn{2}{c}{\textbf{Test Err. (\%)}} & \textbf{Params} & $\times+$ & \textbf{Search Cost} \\
\cmidrule(lr){2-3}
&                            \textbf{top-1} & \textbf{top-5} & \textbf{(M)} & \textbf{(M)} & \textbf{(GPU-days)} \\
\hline
Inception-v1~\cite{szegedy2015going}               & 30.2 & 10.1 & 6.6     & 1448 & -            \\
MobileNet~\cite{howard2017mobilenets}              & 29.4 & 10.5 & 4.2     & 569  & -            \\
ShuffleNet 2$\times$ (v2)~\cite{ma2018shufflenet}  & 25.1 & -    & $\sim$5 & 591  & -            \\
\hline
NASNet-A~\cite{zoph2018learning}                   & 26.0 & 8.4  & 5.3     & 564  & 1800         \\
MnasNet-92~\cite{tan2019mnasnet}                   & 25.2 & 8.0  & 4.4     & 388  & -            \\
AmoebaNet-C~\cite{real2019regularized}             & 24.3 & 7.6  & 6.4     & 570  & 3150         \\
\hline
SNAS (mild)~\cite{xie2018snas}                     & 27.3 & 9.2  & 4.3     & 522  & 1.5          \\
ProxylessNAS$^\ddagger$~\cite{cai2019proxylessnas} & 24.9 & 7.5  & 7.1     & 465  & 8.3          \\
DARTS~\cite{liu2019darts}                          & 26.7 & 8.7  & 4.7     & 574  & 4.0          \\
P-DARTS~\cite{chen2019progressive}                 & 24.4 & 7.4  & 4.9     & 557  & 0.3          \\
PC-DARTS~\cite{xu2020pc}$^\ddagger$                & 24.2 & 7.3  & 5.3     & 597  & 3.8          \\
\hline
GOLD-NAS-I                                         & 24.7 & 7.4  & 5.4     & 586  & 1.1          \\
\hline
GOLD-NAS-X$^\ddagger$                              & 24.3 & 7.3  & 6.4     & 585  & 2.5          \\
GOLD-NAS-Y$^\ddagger$                              & 24.3 & 7.5  & 6.4     & 578  & 2.1          \\
GOLD-NAS-Z$^\ddagger$                              & 24.0 & 7.3  & 6.3     & 585  & 1.7          \\
GOLD-NAS-Z-tr$^\ddagger$                           & 23.9 & 7.3  & 6.4     & 590  & 1.7          \\
\hline
\end{tabular}
\end{center}
\caption{Comparison with state-of-the-arts on ImageNet-1K, under the \textit{mobile setting}. $^\ddagger$: these architectures are searched on ImageNet.}
\label{tab:comparisonImageNet}
\end{wraptable}

Results are summarized in Table~\ref{tab:comparisonImageNet}. GOLD-NAS shows competitive performance among state-of-the-arts. In particular, the transferred GOLD-NAS-I reports a top-1 error of $24.7\%$, and the directly searched GOLD-NAS-Z reports $24.0\%$. Interestingly, GOLD-NAS-Z+ reports $23.9\%$, showing that a longer pruning procedure often leads to higher resource efficiency. Also, GOLD-NAS enjoys smaller search costs, \textit{e.g.}, the cost of GOLD-NAS-Z ($1.7$ GPU-days) is more than $2\times$ faster than prior direct search methods~\cite{cai2019proxylessnas,xu2020pc}.

\begin{wrapfigure}{r}{6.6cm}
\vspace{-1.5cm}
\centering
\begin{subfigure}{6.6cm}
\centering
\includegraphics[width=6.5cm]{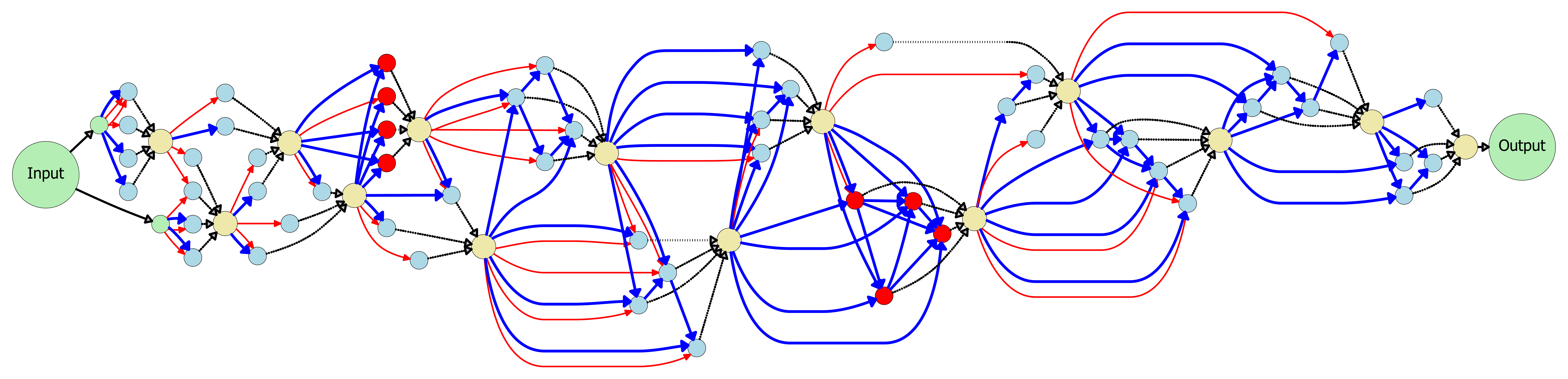}
\caption{GOLD-NAS-X, $24.3\%$ top-1 error}
\end{subfigure}
\begin{subfigure}{6.6cm}
\centering
\includegraphics[width=6.5cm]{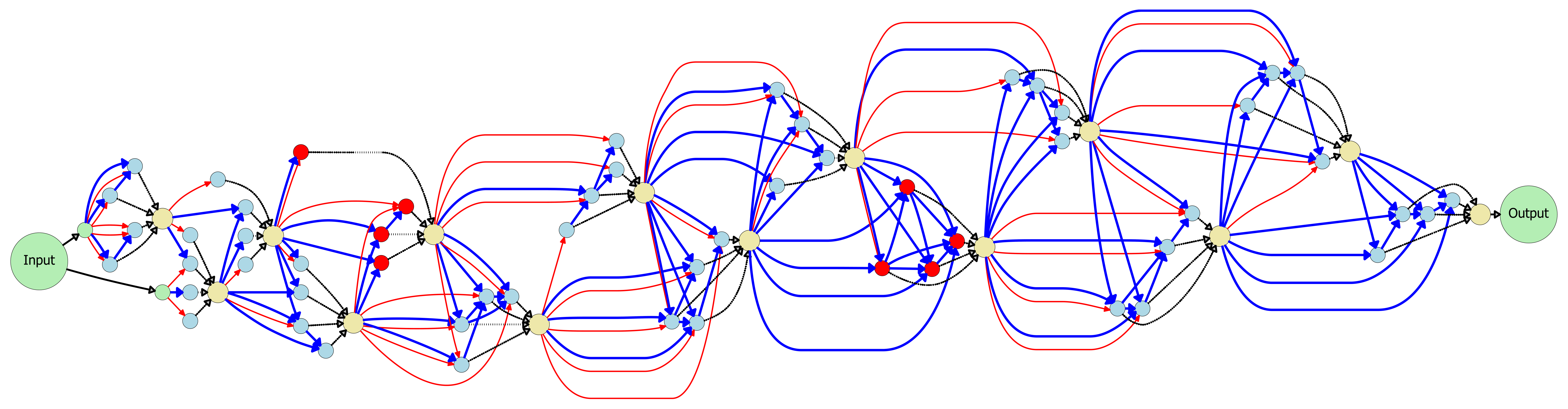}
\caption{GOLD-NAS-Z, $24.0\%$ top-1 error}
\end{subfigure}
\begin{subfigure}{6.6cm}
\centering
\includegraphics[width=6.5cm]{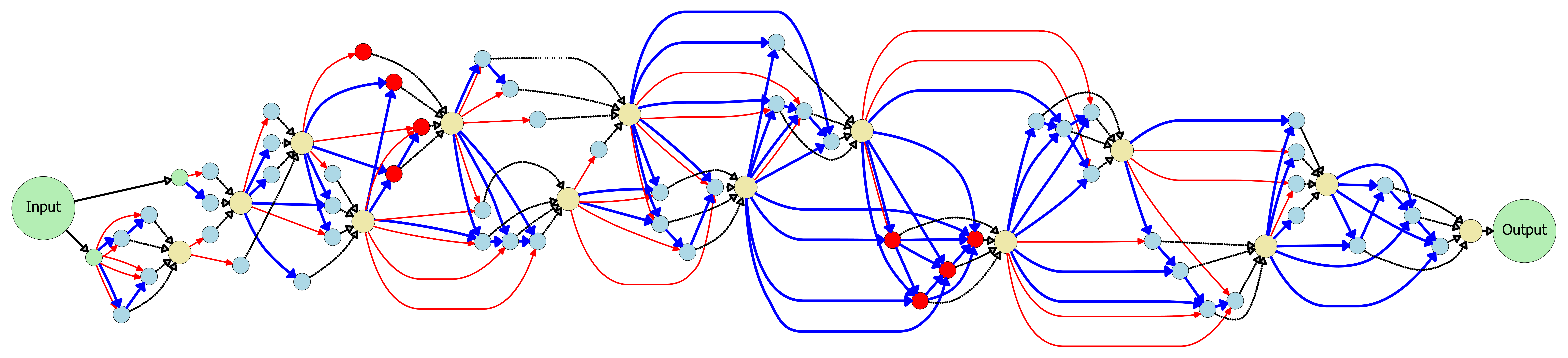}
\caption{GOLD-NAS-Z-tr, $23.9\%$ top-1 error}
\end{subfigure}
\caption{Three architectures found on ImageNet. The \textcolor{red}{red} thin, \textcolor{blue}{blue} thick, and black dashed arrows indicate \textsf{skip-connect}, \textsf{sep-conv-3x3}, and \textsf{concatenation}, respectively. This figure is best viewed in a colored and zoomed-in document.}
\label{fig:visualizationImageNet}
\end{wrapfigure}

The searched architectures on ImageNet are shown in Figure~\ref{fig:visualizationImageNet}. GOLD-NAS tends to increase the portion of \textsf{sep-conv-3x3}, the parameterized operator, in the middle and late stages (close to output) of the network. This leads to an increase of parameters compared to the architectures found in the original space. This implies that GOLD-NAS can assign the computational resource to different network stages more flexibly, which mainly owes to the enlarged search space and the stable search algorithm.

\section{Conclusions}
\label{conclusions}

In this paper, we present a novel algorithm named \textbf{GOLD-NAS} (Gradual One-Level Differentiable Neural Architecture Search). Starting with the need of exploring a more challenging search space, we make use of one-level differentiable optimization and reveal the main reason for the failure lies in the discretization error. To alleviate it, we propose a gradual pruning procedure in which the resource usage plays the role of regularization that increases with time. GOLD-NAS is able to find a set of Pareto-optimal architectures with one search procedure. The search results on CIFAR10 and ImageNet demonstrate that GOLD-NAS achieves a nice tradeoff between accuracy and efficiency.

Our work delivers some new information to the NAS community. \textbf{First}, we encourage the researchers to avoid manually designed rules. This often leads to a larger search space yet very different architectures to be found -- provided with stable search methods, these newly discovered architectures can be more efficient. \textbf{Second} and more importantly, reducing the optimization gap brings benefit to NAS. GOLD-NAS alleviates \textit{discretization error}, one specific type of optimization gap, but it is imperfect as it still requires network reshape and re-training. We are looking forward to extending GOLD-NAS into a completely end-to-end search method that can incorporate various types of hardware constraints. This is an important future research direction.

\section*{Broader Impact}

This paper presents GOLD-NAS, a novel framework for neural architecture search. We summarize the potential impact of our work in the following aspects.
\begin{itemize}
\item \textbf{To the research community.} We advocate for the exploration in an enlarged search space. This is important for the NAS community, because we believe that most existing spaces have incorporated many heuristic rules. Getting rid of these rules not only helps to explore more interesting architectures, but also provides a benchmark for discriminating effective NAS algorithms from manually designed tricks which often fail dramatically in the enlarged space.
\item \textbf{To software-hardware integrated design.} In real-world applications, it is often important to consider the efficiency of a neural network in a specific hardware, \textit{e.g.}, GPU or CPU. GOLD-NAS provides a flexible interface for injecting different kinds of resource constraints. It also enjoys an intriguing ability of producing a Pareto-front during one search procedure. This can save the efforts and computational costs of developers.
\item \textbf{To the downstream engineers.} GOLD-NAS is fast, stable, and easily implemented (with our code released), so it may become a popular choice for engineers to deploy our algorithm to different vision scenarios. While this may help to develop AI-based applications, there exist risks that some engineers, with relatively less knowledge in deep learning, can deliberately use the algorithm, \textit{e.g.}, without considering the amount of training data, which may actually harm the performance of the designed system.
\item \textbf{To the society.} There is a long-lasting debate on the impact that AI can bring to the human society. Being an algorithm for improving the fundamental ability of deep learning, our work lies on the path of advancing AI. Therefore, in general, it can bring both beneficial and harmful impacts and it really depends on the motivation of the users.
\end{itemize}

We also encourage the community to investigate the following problems.
\begin{enumerate}
\item Is the current search space sufficiently complex? Is it possible to challenge the NAS algorithms with even more difficult search spaces?
\item If there exist multiple resource constraints (\textit{e.g.}, model size, latency, \textit{etc.}), how to schedule the regularization coefficients in the gradual pruning process?
\item Is it possible to deploy the search procedure to other vision tasks (\textit{e.g.}, detection, segmentation, \textit{etc.}) or unsupervised representation learning?
\end{enumerate}

{\small
\bibliographystyle{abbrv}
\bibliography{refs}

\begin{thebibliography}{10}

\bibitem{bi2019stabilizing}
K.~Bi, C.~Hu, L.~Xie, X.~Chen, L.~Wei, and Q.~Tian.
\newblock Stabilizing darts with amended gradient estimation on architectural
  parameters.
\newblock {\em arXiv preprint arXiv:1910.11831}, 2019.

\bibitem{cai2018efficient}
H.~Cai, T.~Chen, W.~Zhang, Y.~Yu, and J.~Wang.
\newblock Efficient architecture search by network transformation.
\newblock In {\em AAAI Conference on Artificial Intelligence}, 2018.

\bibitem{cai2019proxylessnas}
H.~Cai, L.~Zhu, and S.~Han.
\newblock Proxylessnas: Direct neural architecture search on target task and
  hardware.
\newblock 2019.

\bibitem{chen2019progressive}
X.~Chen, L.~Xie, J.~Wu, and Q.~Tian.
\newblock Progressive differentiable architecture search: Bridging the depth
  gap between search and evaluation.
\newblock In {\em International Conference on Computer Vision}, 2019.

\bibitem{chu2019fairnas}
X.~Chu, B.~Zhang, R.~Xu, and J.~Li.
\newblock Fairnas: Rethinking evaluation fairness of weight sharing neural
  architecture search.
\newblock {\em arXiv preprint arXiv:1907.01845}, 2019.

\bibitem{chu2019fair}
X.~Chu, T.~Zhou, B.~Zhang, and J.~Li.
\newblock Fair darts: Eliminating unfair advantages in differentiable
  architecture search.
\newblock {\em arXiv preprint arXiv:1911.12126}, 2019.

\bibitem{cubuk2019autoaugment}
E.~D. Cubuk, B.~Zoph, D.~Mane, V.~Vasudevan, and Q.~V. Le.
\newblock Autoaugment: Learning augmentation policies from data.
\newblock In {\em Computer Vision and Pattern Recognition}, 2019.

\bibitem{deng2009imagenet}
J.~Deng, W.~Dong, R.~Socher, L.~J. Li, K.~Li, and L.~Fei-Fei.
\newblock Imagenet: A large-scale hierarchical image database.
\newblock In {\em Computer Vision and Pattern Recognition}, 2009.

\bibitem{devries2017improved}
T.~DeVries and G.~W. Taylor.
\newblock Improved regularization of convolutional neural networks with cutout.
\newblock {\em arXiv preprint arXiv:1708.04552}, 2017.

\bibitem{guo2019single}
Z.~Guo, X.~Zhang, H.~Mu, W.~Heng, Z.~Liu, Y.~Wei, and J.~Sun.
\newblock Single path one-shot neural architecture search with uniform
  sampling.
\newblock {\em arXiv preprint arXiv:1904.00420}, 2019.

\bibitem{he2016deep}
K.~He, X.~Zhang, S.~Ren, and J.~Sun.
\newblock Deep residual learning for image recognition.
\newblock In {\em Computer Vision and Pattern Recognition}, 2016.

\bibitem{howard2019searching}
A.~Howard, M.~Sandler, G.~Chu, L.-C. Chen, B.~Chen, M.~Tan, W.~Wang, Y.~Zhu,
  R.~Pang, V.~Vasudevan, et~al.
\newblock Searching for mobilenetv3.
\newblock In {\em International Conference on Computer Vision}, 2019.

\bibitem{howard2017mobilenets}
A.~G. Howard, M.~Zhu, B.~Chen, D.~Kalenichenko, W.~Wang, T.~Weyand,
  M.~Andreetto, and H.~Adam.
\newblock Mobilenets: Efficient convolutional neural networks for mobile vision
  applications.
\newblock {\em arXiv preprint arXiv:1704.04861}, 2017.

\bibitem{huang2017densely}
G.~Huang, Z.~Liu, L.~Van Der~Maaten, and K.~Q. Weinberger.
\newblock Densely connected convolutional networks.
\newblock In {\em Computer Vision and Pattern Recognition}, 2017.

\bibitem{krizhevsky2009learning}
A.~Krizhevsky and G.~Hinton.
\newblock Learning multiple layers of features from tiny images.
\newblock Technical report, 2009.

\bibitem{krizhevsky2012imagenet}
A.~Krizhevsky, I.~Sutskever, and G.~E. Hinton.
\newblock Imagenet classification with deep convolutional neural networks.
\newblock In {\em Advances in Neural Information Processing Systems}, 2012.

\bibitem{lecun2015deep}
Y.~LeCun, Y.~Bengio, and G.~Hinton.
\newblock Deep learning.
\newblock {\em Nature}, 521(7553):436--444, 2015.

\bibitem{li2019stacnas}
G.~Li, X.~Zhang, Z.~Wang, Z.~Li, and T.~Zhang.
\newblock Stacnas: Towards stable and consistent optimization for
  differentiable neural architecture search.
\newblock {\em arXiv preprint arXiv:1909.11926}, 2019.

\bibitem{li2019random}
L.~Li and A.~Talwalkar.
\newblock Random search and reproducibility for neural architecture search.
\newblock {\em arXiv preprint arXiv:1902.07638}, 2019.

\bibitem{liang2019darts+}
H.~Liang, S.~Zhang, J.~Sun, X.~He, W.~Huang, K.~Zhuang, and Z.~Li.
\newblock Darts+: Improved differentiable architecture search with early
  stopping.
\newblock {\em arXiv preprint arXiv:1909.06035}, 2019.

\bibitem{liu2018progressive}
C.~Liu, B.~Zoph, M.~Neumann, J.~Shlens, W.~Hua, L.~J. Li, L.~Fei-Fei, A.~L.
  Yuille, J.~Huang, and K.~Murphy.
\newblock Progressive neural architecture search.
\newblock In {\em European Conference on Computer Vision}, 2018.

\bibitem{liu2019darts}
H.~Liu, K.~Simonyan, and Y.~Yang.
\newblock Darts: Differentiable architecture search.
\newblock In {\em International Conference on Learning Representations}, 2019.

\bibitem{luo2018neural}
R.~Luo, F.~Tian, T.~Qin, E.~Chen, and T.~Y. Liu.
\newblock Neural architecture optimization.
\newblock In {\em Advances in Neural Information Processing Systems}, 2018.

\bibitem{ma2018shufflenet}
N.~Ma, X.~Zhang, H.~T. Zheng, and J.~Sun.
\newblock Shufflenet v2: Practical guidelines for efficient cnn architecture
  design.
\newblock In {\em European Conference on Computer Vision}, 2018.

\bibitem{nayman2019xnas}
N.~Nayman, A.~Noy, T.~Ridnik, I.~Friedman, R.~Jin, and L.~Zelnik-Manor.
\newblock Xnas: Neural architecture search with expert advice.
\newblock {\em arXiv preprint arXiv:1906.08031}, 2019.

\bibitem{pham2018efficient}
H.~Pham, M.~Guan, B.~Zoph, Q.~V. Le, and J.~Dean.
\newblock Efficient neural architecture search via parameter sharing.
\newblock In {\em International Conference on Machine Learning}, 2018.

\bibitem{real2019regularized}
E.~Real, A.~Aggarwal, Y.~Huang, and Q.~V. Le.
\newblock Regularized evolution for image classifier architecture search.
\newblock In {\em AAAI Conference on Artificial Intelligence}, 2019.

\bibitem{real2017large}
E.~Real, S.~Moore, A.~Selle, S.~Saxena, Y.~L. Suematsu, J.~Tan, Q.~V. Le, and
  A.~Kurakin.
\newblock Large-scale evolution of image classifiers.
\newblock In {\em International Conference on Machine Learning}, 2017.

\bibitem{russakovsky2015imagenet}
O.~Russakovsky, J.~Deng, H.~Su, J.~Krause, S.~Satheesh, S.~Ma, Z.~Huang,
  A.~Karpathy, A.~Khosla, M.~Bernstein, et~al.
\newblock Imagenet large scale visual recognition challenge.
\newblock {\em International journal of computer vision}, 115(3):211--252,
  2015.

\bibitem{shin2018differentiable}
R.~Shin, C.~Packer, and D.~Song.
\newblock Differentiable neural network architecture search.
\newblock In {\em International Conference on Learning Representations --
  Workshops}, 2018.

\bibitem{simonyan2015very}
K.~Simonyan and A.~Zisserman.
\newblock Very deep convolutional networks for large-scale image recognition.
\newblock In {\em International Conference on Learning Representations}, 2015.

\bibitem{szegedy2015going}
C.~Szegedy, W.~Liu, Y.~Jia, P.~Sermanet, S.~Reed, D.~Anguelov, D.~Erhan,
  V.~Vanhoucke, and A.~Rabinovich.
\newblock Going deeper with convolutions.
\newblock In {\em Computer Vision and Pattern Recognition}, 2015.

\bibitem{tan2019mnasnet}
M.~Tan, B.~Chen, R.~Pang, V.~Vasudevan, M.~Sandler, A.~Howard, and Q.~V. Le.
\newblock Mnasnet: Platform-aware neural architecture search for mobile.
\newblock In {\em Computer Vision and Pattern Recognition}, 2019.

\bibitem{wu2019fbnet}
B.~Wu, X.~Dai, P.~Zhang, Y.~Wang, F.~Sun, Y.~Wu, Y.~Tian, P.~Vajda, Y.~Jia, and
  K.~Keutzer.
\newblock Fbnet: Hardware-aware efficient convnet design via differentiable
  neural architecture search.
\newblock In {\em Computer Vision and Pattern Recognition}, pages 10734--10742,
  2019.

\bibitem{xie2017genetic}
L.~Xie and A.~L. Yuille.
\newblock Genetic {CNN}.
\newblock In {\em International Conference on Computer Vision}, 2017.

\bibitem{xie2018snas}
S.~Xie, H.~Zheng, C.~Liu, and L.~Lin.
\newblock {SNAS}: Stochastic neural architecture search.
\newblock {\em arXiv preprint arXiv:1812.09926}, 2018.

\bibitem{xu2020pc}
Y.~Xu, L.~Xie, X.~Zhang, X.~Chen, G.~J. Qi, Q.~Tian, and H.~Xiong.
\newblock Pc-darts: Partial channel connections for memory-efficient
  differentiable architecture search.
\newblock In {\em International Conference on Learning Representations}, 2020.

\bibitem{xu2020latency}
Y.~Xu, L.~Xie, X.~Zhang, X.~Chen, B.~Shi, Q.~Tian, and H.~Xiong.
\newblock Latency-aware differentiable neural architecture search.
\newblock {\em arXiv preprint arXiv:2001.06392}, 2020.

\bibitem{zela2020understanding}
A.~Zela, T.~Elsken, T.~Saikia, Y.~Marrakchi, T.~Brox, and F.~Hutter.
\newblock Understanding and robustifying differentiable architecture search.
\newblock In {\em International Conference on Learning Representations}, 2020.

\bibitem{zoph2017neural}
B.~Zoph and Q.~V. Le.
\newblock Neural architecture search with reinforcement learning.
\newblock In {\em International Conference on Learning Representations}, 2017.

\bibitem{zoph2018learning}
B.~Zoph, V.~Vasudevan, J.~Shlens, and Q.~V. Le.
\newblock Learning transferable architectures for scalable image recognition.
\newblock In {\em Computer Vision and Pattern Recognition}, 2018.

\end{thebibliography}
}

\appendix

\section{Details of the Search Algorithm}
\label{details}

\subsection{How Many Architectures Are There in the New Search Space?}
\label{details:space}

In the enlarged search space, each cell (either normal or reduction) is individually determined. In each cell, there are $4$ intermediate nodes, each of which can receive inputs from its precedents. In each edge, there are two possible operators, \textsf{sep-conv-3x3} and \textsf{skip-connect}. To guarantee validity, each node must have at least one preserved operator (on any edge). This means that the $n$-th node (${n}={2,3,4,5}$) contributes $2^{2n}-1$ possibilities because each of the $2n$ operators can be on or off, but the situation that all operators are off is invalid. Therefore, there are ${\left(2^4-1\right)\left(2^6-1\right)\left(2^8-1\right)\left(2^{10}-1\right)}\approx{2.5\times10^8}$ combinations for each cell.

There are $14$ or $20$ cells, according to the definition of DARTS. If $14$ cells are used, the total number of architectures in the search space is ${\left(2.5\times10^8\right)^{14}}\approx{3.1\times10^{117}}$; if $20$ cells are used, the number becomes ${\left(2.5\times10^8\right)^{20}}\approx{6.9\times10^{167}}$. Of course, under a specific FLOPs constraint, the number of architectures is much smaller than this number, but our space is still much more complex than the original one -- this is a side factor that we can find a series of Pareto-optimal architectures in one search procedure.

\subsection{One-Level Search in the Original Search Space}
\label{details:onelevel}

DARTS reported that one-level optimization failed dramatically in the original search space, \textit{i.e.}, the test error is $3.56\%$ on CIFAR10, which is even inferior to random search ($3.29\pm0.15\%$). We reproduced one-level optimization and reported a similar error rate of $3.54\%$.

We find that the failure is mostly caused by the over-fitting issue, as we have explained in the main article: the number of network weights is much larger than the number of architectural parameters, so optimizing the former is more effective but delivers no information to architecture search. To alleviate this issue, we add data augmentation to the original one-level optimization (only in the search phase, the re-training phase is unchanged at all). With merely this simple modification, the one-level searched architecture reports an error rate of $2.80\pm0.06\%$, which is comparable to the second-order optimization of DARTS and outperforms the first-order optimization of DARTS -- note that both first-order and second-order optimization needs bi-level optimization. This verifies the potential of one-level optimization -- more importantly, one-level optimization gets rid of the burden of inaccurate gradient estimation of bi-level optimization.

\subsection{Calculation of the FLOPs Function}
\label{details:flops}

We first elaborate the ideology of designing the function. For a specific operator $o$, its FLOPs is easily measured by some mathematical calculation and written as a function of $\mathrm{FLOPs}\!\left(o\right)$. When we consider the architectural parameter $\boldsymbol{\alpha}$ in a differentiable search procedure, we should notice that the FLOPs term, $\mathrm{FLOPs}\!\left(\boldsymbol{\alpha}\right)$, reflects the \textit{expectation} of the FLOPs of the current architecture (parameterized by $\boldsymbol{\alpha}$). The calculation of $\mathrm{FLOPs}\!\left(\boldsymbol{\alpha}\right)$ should consider three key points. For each individual operator $o$ with an architectural parameter of $\alpha$, (i) its expected FLOPs should increase with $\alpha$, in particular, $\sigma\!\left(\alpha\right)$; (ii) to remove an operator from an edge, the average $\sigma\!\left(\alpha\right)$ value in the edge should be considered; (iii) as $\sigma\!\left(\alpha\right)$ goes towards $1$, the penalty that it receives should increase slower -- this is to facilitate a concentration of weights. Considering the above condition, we design the FLOPs function as follows:
\begin{equation}
{\mathrm{FLOPs}\!\left(\boldsymbol{\alpha}\right)}={\sum_{o}\ln\!\left(1+\sigma\!\left(\alpha_o\right)/\overline{\sigma\!\left(\boldsymbol{\alpha}\right)}\right)\cdot\mathrm{FLOPs}\!\left(o\right)},
\end{equation}
where the design of $\ln\!\left(1+\cdot\right)$ is to guarantee convexity, and we believe this form is not the optimal choice. The \textit{uniform} version of $\mathrm{FLOPs}\!\left(\boldsymbol{\alpha}\right)$, $\overline{\mathrm{FLOPs}}\!\left(\boldsymbol{\alpha}\right)$, is computed via setting $\mathrm{FLOPs}(o)$ of all operators to be identical, so that the search algorithm mainly focuses on the impact of each operator on the classification error. That is to say, $\overline{\mathrm{FLOPs}}\!\left(\boldsymbol{\alpha}\right)$ suppresses the operator with the least contribution to the classification task while $\mathrm{FLOPs}(\boldsymbol{\alpha})$ tends to suppress the most expensive operator first.

In practice, we use the \texttt{thop} library to calculate the terms of $\mathrm{FLOPs}\!\left(o\right)$. Let $C$ be the number of input and output channels, and $W$ and $H$ be the width and height of the output. Then, the FLOPs of a \textsf{skip-connect} operator is $0$ if the stride is $1$ and ${\mathrm{FLOPs}\!\left(o\right)}={C^2HW}$ if the stride is $2$, and the FLOPs of a \textsf{sep-conv-3x3} operator is ${\mathrm{FLOPs}\!\left(o\right)}=2\times\left(C^2HW+9\times{}CHW\right)$ (note that there are two cascaded convolutions in this operator).

\section{Full Results of CIFAR10 Experiments}
\label{cifar_}

\subsection{Hyper-parameter Settings}
\label{cifar_:parameters}

First of all, we tried both $14$-cell and $20$-cell settings for the entire network, and found that they produced similar performance but the $14$-cell setting is more efficient, so we keep this setting throughout the remaining part of this paper.

We use $14$ cells for both search and re-train procedure, and the initial channels before the first cell is set to be $36$. During the search procedure, all the architectural parameters are initialized to zero. The batch size is set to be $96$. An SGD optimizer with a momentum of $0.9$ is used to update the architectural parameters, $\boldsymbol{\alpha}$, and the learning rate $\eta_{\boldsymbol{\alpha}}$ is set to be $1$. Another SGD optimizer is used to update the network parameters, and the only difference is that the learning rate $\eta_{\boldsymbol{\omega}}$ is set to be $0.01$. The pruning pace $n_0$ is set to be 4, and it could be either increased to accelerate the search process (faster but less accurate) or decreased to smooth the search process (slower but more accurate). The pruning thresholds $\xi_{max}$ and $\xi_{min}$ are set to be $0.05$ and $0.01$. $c_0$ is set to be $2$ for simplicity, and similar to $n_0$, it can be adjusted to change the pace of the pruning process. $\lambda_0$ is set to be $1\times10^{-5}$, which is chosen to make the two terms of loss function comparable to each other. $t_0$ is set to be $3$, and it can be increased to improve the stability of the Pareto-optimal architectures or decreased to obtain a larger number of Pareto-optimal architectures. $\mathrm{FLOPs}_\mathrm{min}$ is set to be $240\mathrm{M}$ for ${\mu}={1}$ and $360\mathrm{M}$ for ${\mu}={0}$: this parameter is not very important because we can terminate the search process at anywhere we want. AutoAugment is applied in the search procedure to avoid over-fitting (\textit{i.e.}, the network is easily biased towards tuning $\boldsymbol{\omega}$ than $\boldsymbol{\alpha}$, see Appendix~\ref{details:onelevel}), but we do not use it during the re-training process for the fair comparison against existing approaches.

\begin{wrapfigure}{r}{6cm}
\vspace{-0.5cm}
\includegraphics[width=6cm]{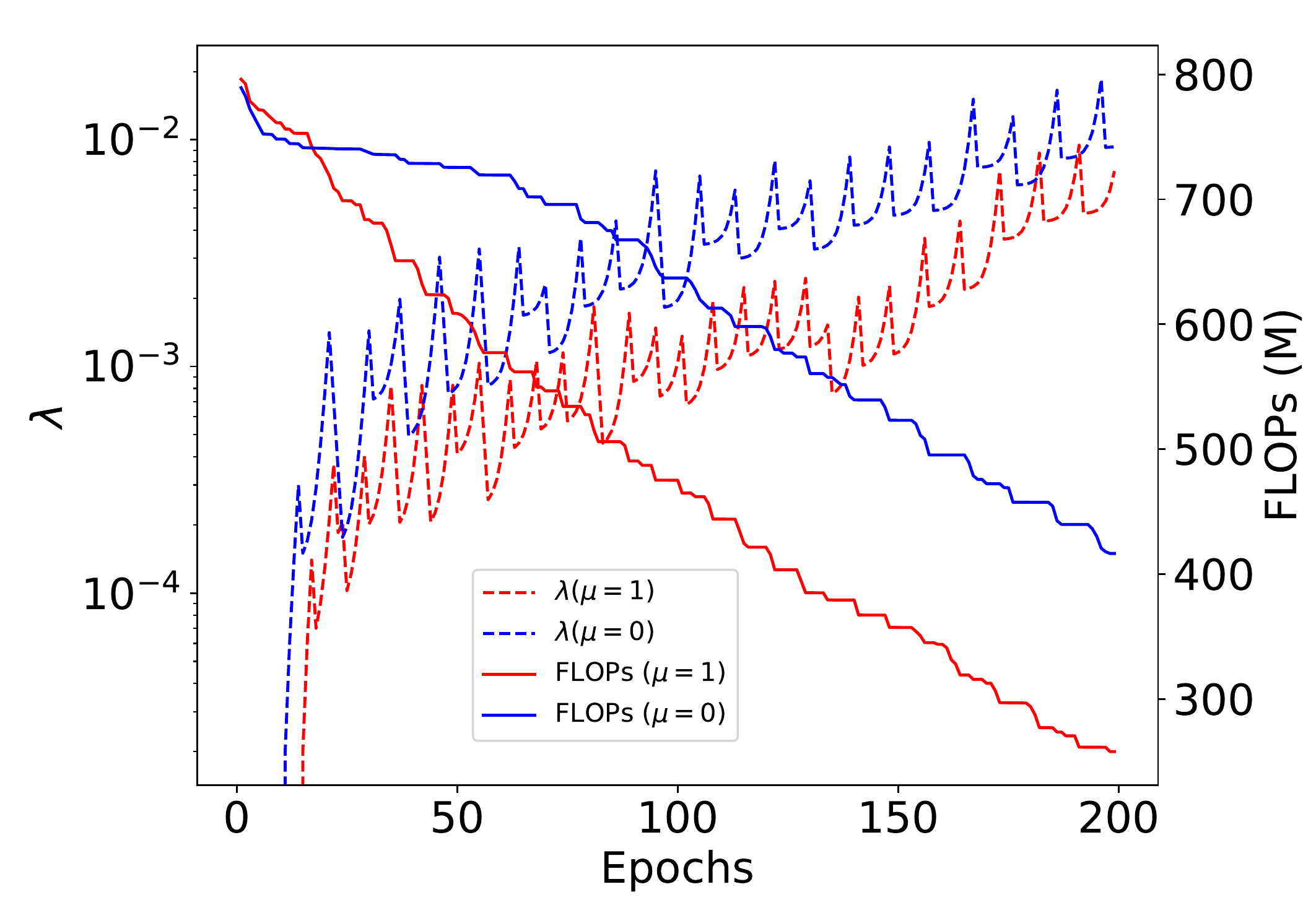}
\caption{Visualization of two search procedures on CIFAR10 with ${\eta}={0}$ (\textcolor{blue}{blue}) and ${\eta}={1}$ (\textcolor{red}{red}), respectively. $\lambda$ zigzags from $0$ to a large value, and the FLOPs of the super-network goes down.}
\label{fig:procedure}
\vspace{-0.95cm}
\end{wrapfigure}

The re-training process remains the same as the convention. Each Pareto-optimal architecture is trained for $600$ epochs with a batch size of $96$. We use SGD optimizer with a momentum of $0.9$, and the corresponding learning rate is initialized to $0.025$ and annealed to zero following a cosine schedule. We use cutout, path Dropout with a probability of $0.2$, and an auxiliary tower with a weight of $0.4$ during the training process. The training process takes $0.3$ to $1.2$ days on a single NVIDIA Tesla-V100 GPU, according to the complexity (FLOPs) of each search architecture.

\subsection{Analysis on the Search Procedure}
\label{cifar_:procedure}

In Figure~\ref{fig:procedure}, we visualize the search procedures on CIFAR10 using the hyper-parameters of ${\eta}={0}$ and ${\eta}={1}$. We can observe that, as the search procedure goes, weak operators are pruned out from the super-network and the FLOPs of the network gradually goes down. With ${\eta}={1}$, the rate of pruning is much faster. More interestingly, $\lambda$, the balancing coefficient, zigzags from a small value to a large value. In each period, $\lambda$ first goes up to force some operators to have lower weights (during this process, nothing is pruned and the architecture remains unchanged), and then goes down as pruning takes effect to eliminate the weak operators. Each local maximum (just before the pruning stage) corresponds to a Pareto-optimal architecture.

\subsection{Visualization of the Searched Architectures}
\label{cifar_:visualization}

\begin{figure}[!t]
\centering
\begin{subfigure}{6.6cm}
\centering
\stackunder[0.2cm]{\includegraphics[width=6.5cm]{architecture-A.pdf}}{GOLD-NAS-A, $1.58\mathrm{M}$, $2.99\%$ error}
\end{subfigure}
\hfill
\begin{subfigure}{6.6cm}
\centering
\stackunder[0.2cm]{\includegraphics[width=6.5cm]{architecture-G.pdf}}{GOLD-NAS-G, $2.22\mathrm{M}$, $2.75\%$ error}
\end{subfigure}
\begin{subfigure}{6.6cm}
\centering
\stackunder[0.2cm]{\includegraphics[width=6.5cm]{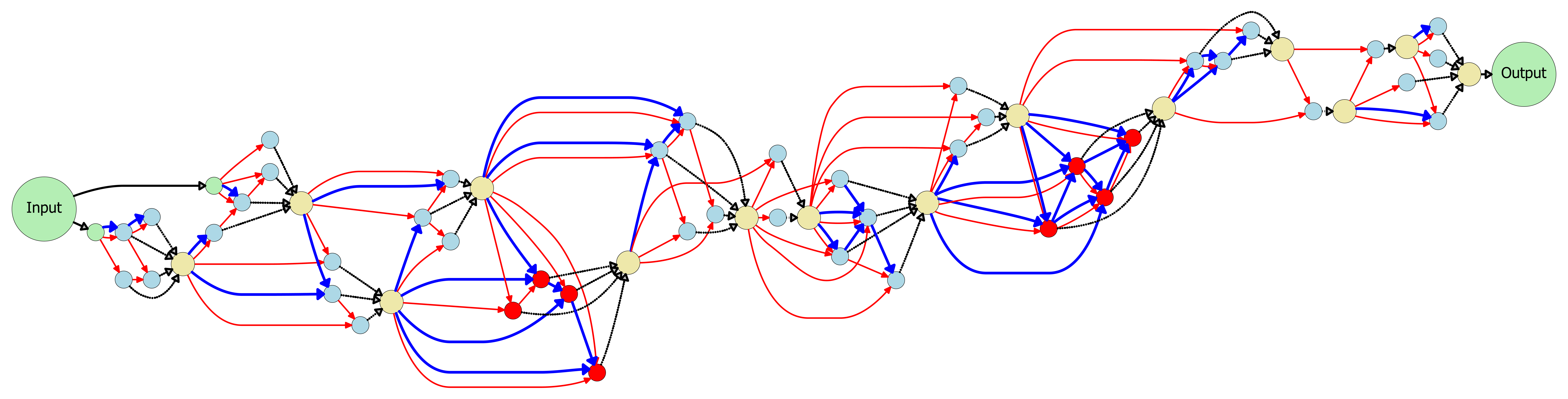}}{GOLD-NAS-B, $1.72\mathrm{M}$, $2.97\%$ error}
\end{subfigure}
\hfill
\begin{subfigure}{6.6cm}
\centering
\stackunder[0.2cm]{\includegraphics[width=6.5cm]{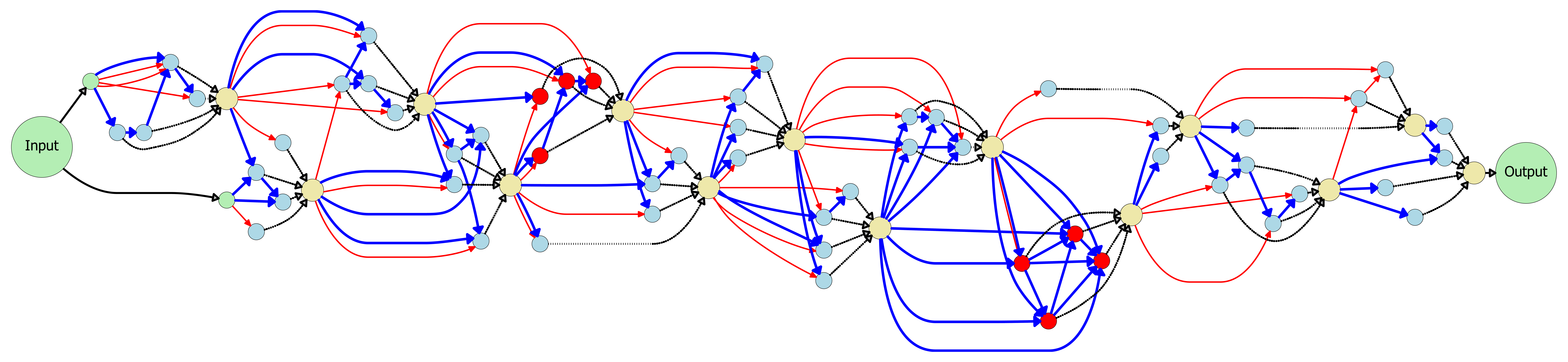}}{GOLD-NAS-H, $2.51\mathrm{M}$, $2.70\%$ error}
\end{subfigure}
\begin{subfigure}{6.6cm}
\centering
\stackunder[0.2cm]{\includegraphics[width=6.5cm]{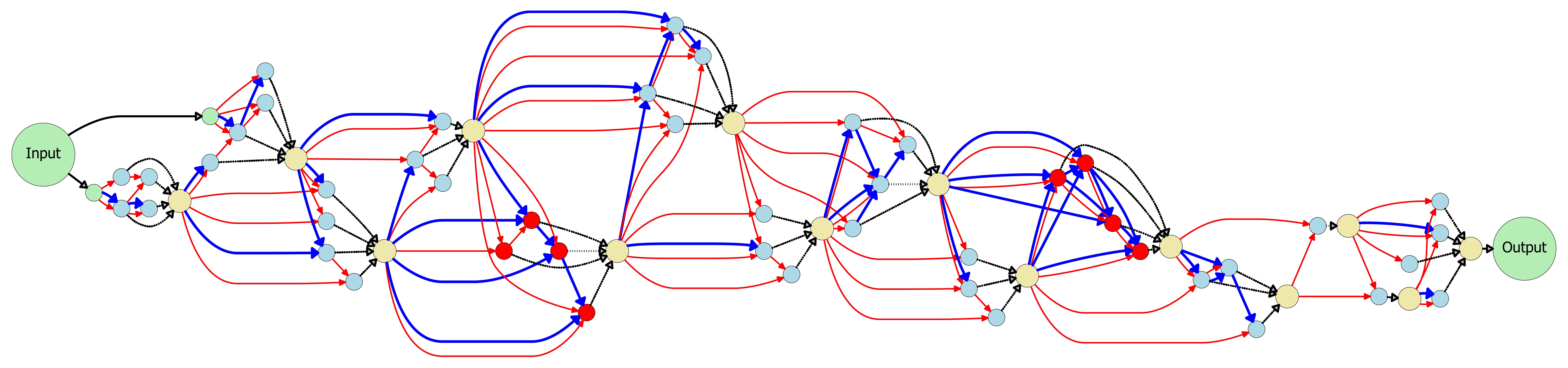}}{GOLD-NAS-C, $1.76\mathrm{M}$, $2.96\%$ error}
\end{subfigure}
\hfill
\begin{subfigure}{6.6cm}
\centering
\stackunder[0.2cm]{\includegraphics[width=6.5cm]{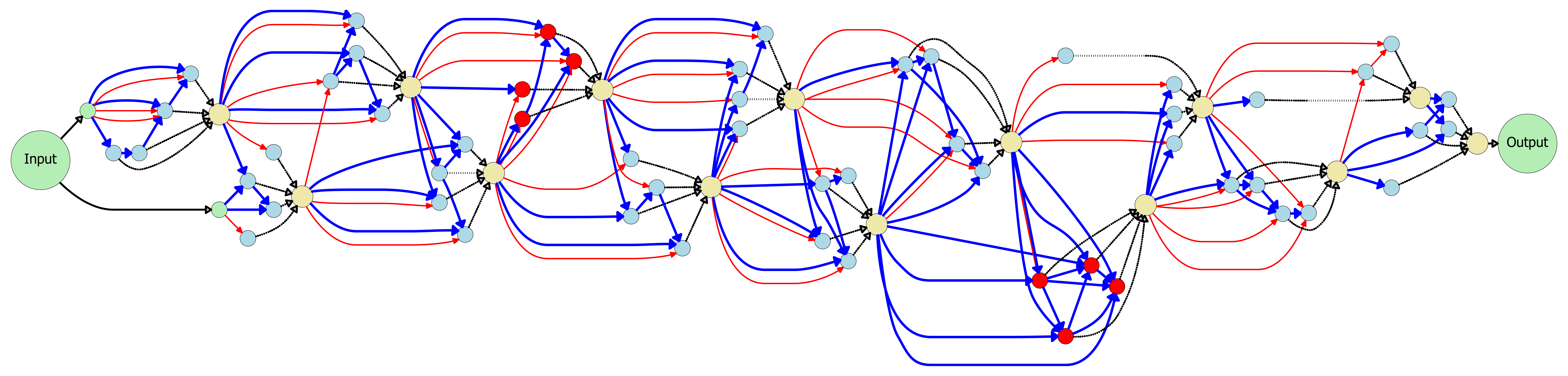}}{GOLD-NAS-I, $2.85\mathrm{M}$, $2.61\%$ error}
\end{subfigure}
\begin{subfigure}{6.6cm}
\centering
\stackunder[0.2cm]{\includegraphics[width=6.5cm]{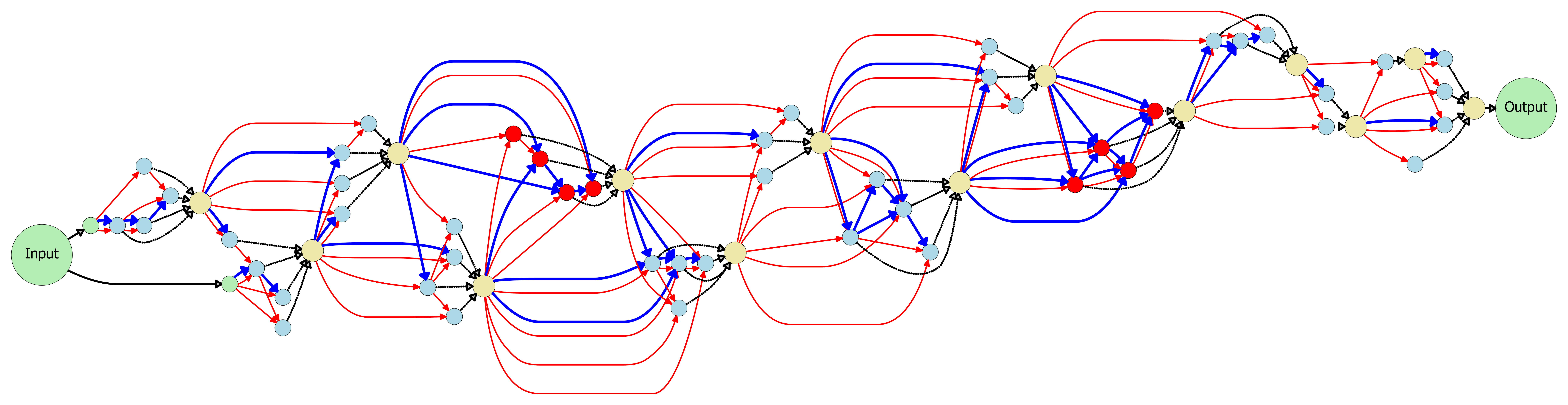}}{GOLD-NAS-D, $1.89\mathrm{M}$, $2.90\%$ error}
\end{subfigure}
\hfill
\begin{subfigure}{6.6cm}
\centering
\stackunder[0.2cm]{\includegraphics[width=6.5cm]{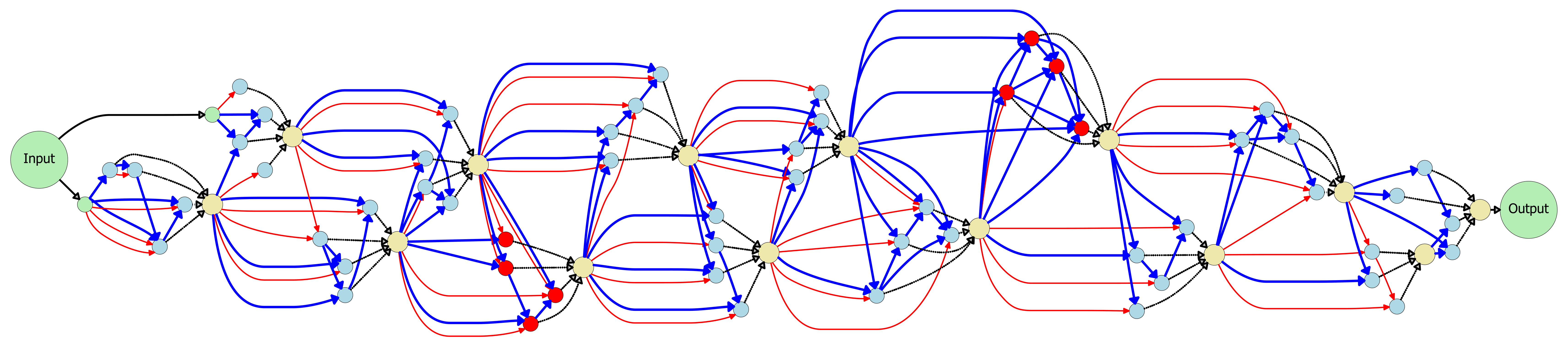}}{GOLD-NAS-J, $3.01\mathrm{M}$, $2.60\%$ error}
\end{subfigure}
\begin{subfigure}{6.6cm}
\centering
\stackunder[0.2cm]{\includegraphics[width=6.5cm]{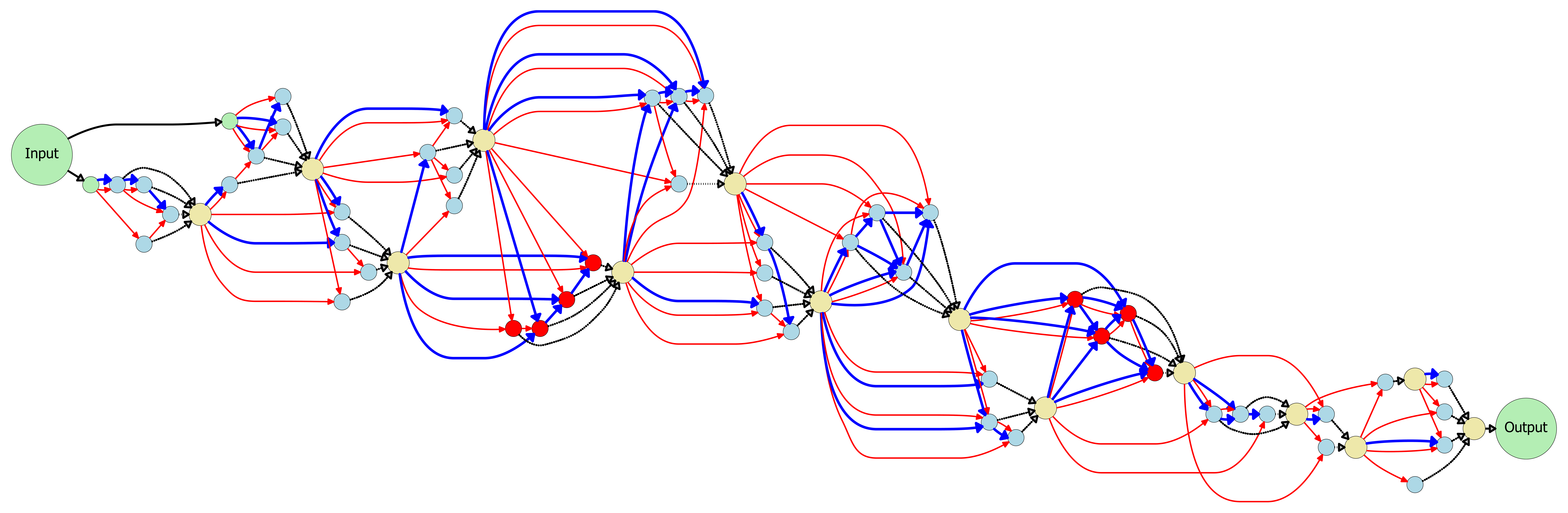}}{GOLD-NAS-E, $1.99\mathrm{M}$, $2.83\%$ error}
\end{subfigure}
\hfill
\begin{subfigure}{6.6cm}
\centering
\stackunder[0.2cm]{\includegraphics[width=6.5cm]{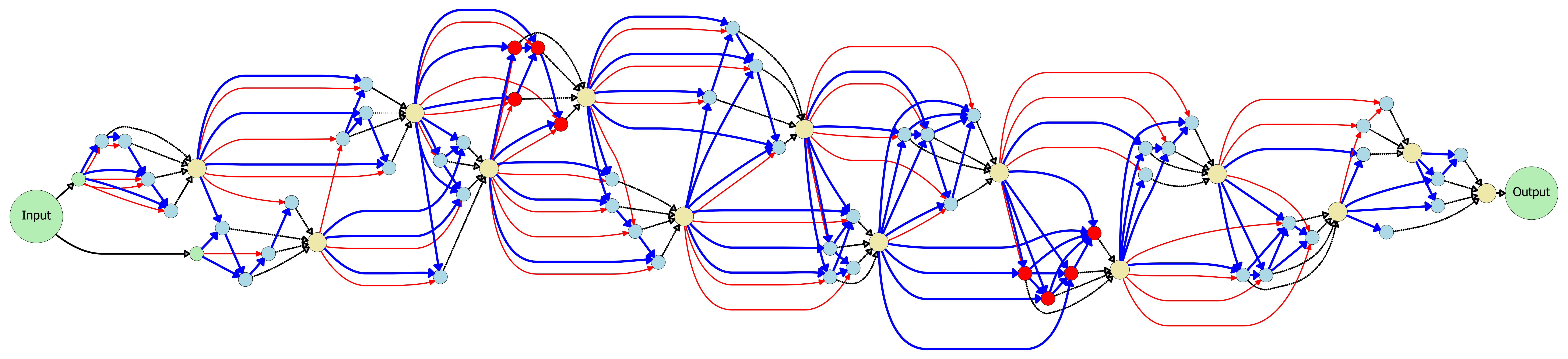}}{GOLD-NAS-K, $3.30\mathrm{M}$, $2.57\%$ error}
\end{subfigure}
\begin{subfigure}{6.6cm}
\centering
\stackunder[0.2cm]{\includegraphics[width=6.5cm]{architecture-F.pdf}}{GOLD-NAS-F, $2.08\mathrm{M}$, $2.81\%$ error}
\end{subfigure}
\hfill
\begin{subfigure}{6.6cm}
\centering
\stackunder[0.2cm]{\includegraphics[width=6.5cm]{architecture-L.pdf}}{GOLD-NAS-L, $3.67\mathrm{M}$, $2.53\%$ error}
\end{subfigure}
\caption{All architectures searched on CIFAR10 during two pruning procedures, ${\eta}={1}$ on the left side, ${\eta}={0}$ on the right side. The \textcolor{red}{red} thin, \textcolor{blue}{blue} thick, and black dashed arrows indicate \textsf{skip-connect}, \textsf{sep-conv-3x3}, and \textsf{concatenation}, respectively. This figure is best viewed in a colored and zoomed-in document.}
\label{fig:visualizationC10all}
\end{figure}

We show all searched architectures on CIFAR10 in Figure~\ref{fig:visualizationC10all}.

\subsection{Details of Random Search Experiments}
\label{cifar_:random}

To produce the random search baseline, we randomly prune out operators from the super-network until the architecture fits the hardware constraint (\textit{e.g.}, FLOPs). It is possible that the architecture becomes invalid during the random pruning process, and we discard such architectures. Each random search process collects $24$ architectures and we train each of them for $100$ epochs and pick up the best one for an entire $600$-epoch re-training. As reported in the paper, we perform random search three times and the best architecture reports an average accuracy of $3.31\pm0.50\%$.

\section{Full Results of ImageNet Experiments}
\label{imagenet_}

\subsection{Hyper-parameter Settings}
\label{imagenet_:parameters}

Following FBNet~\cite{wu2019fbnet} and PC-DARTS~\cite{xu2020pc}, we randomly sample $100$ classes from the original $1\rm{,}000$ classes of ImageNet to reduce the search cost. We do not AutoAugment during the search procedure as the training set is sufficiently large to avoid over-fitting. Other super-parameters are kept unchanged as the CIFAR10 experiments except for $\mathrm{FLOPs}_\mathrm{min}$, which is set to be $500\mathrm{M}$ for the ImageNet experiments.

During the re-training process, the total number of epochs is set to be $250$. The batch size is set to be $1\rm{,}024$ (eight cards). We use an SGD optimizer with an initial learning rate of $0.5$ (decayed linearly after each epoch till $0$), a momentum of $0.9$ and a weight decay of $3\times10^{-5}$. The search process takes around $3$ days on eight NVIDIA Telsa-V100 GPUs.

\end{document}